\documentclass[10pt,journal,compsoc]{IEEEtran}

\ifCLASSOPTIONcompsoc
\usepackage[utf8]{inputenc} 
\usepackage[T1]{fontenc}    
\usepackage{hyperref}
\usepackage{url}  
\usepackage{booktabs}
\usepackage{amsfonts}  
\usepackage{nicefrac} 
\usepackage{microtype}   
\usepackage{fancyhdr}      
\usepackage{bm}
\usepackage{amsmath}
\usepackage{algorithmic}
\usepackage{algorithm}
\usepackage{multirow}
\usepackage{graphicx}
\usepackage{subfigure}
\usepackage{wrapfig}
\usepackage{amsthm}
\usepackage{multicol}
\usepackage{amssymb}
\usepackage{pdfpages}
\usepackage{makecell}
\usepackage{xspace}
\usepackage{float}
\usepackage{threeparttable}
\usepackage{soul}

\makeatletter
\DeclareRobustCommand\onedot{\futurelet\@let@token\@onedot}
\def\@onedot{\ifx\@let@token.\else.\null\fi\xspace}

\def\eg{\emph{e.g}\onedot} 
\def\ie{\emph{i.e}\onedot} 
 
 \def\vs{\emph{vs}\onedot}
 
\def\etal{\emph{et al}\onedot}
\makeatother

\allowdisplaybreaks[4]

\pdfoutput=1

\usepackage[nocompress]{cite}
\else
  \usepackage{cite}
\fi

\ifCLASSINFOpdf
\else
\fi

\begin{document}

\title{Audio-Visual Segmentation with Semantics}

\author{Jinxing Zhou\IEEEauthorrefmark{1},
        Xuyang Shen\IEEEauthorrefmark{1},
        Jianyuan Wang\IEEEauthorrefmark{1},
        Jiayi Zhang,
        Weixuan Sun,
        Jing Zhang,\\
        Stan Birchfield,
        Dan Guo,
        Lingpeng Kong,
        Meng Wang\IEEEauthorrefmark{2},~\IEEEmembership{Fellow,~IEEE},
        and Yiran Zhong\IEEEauthorrefmark{2}

\IEEEcompsocitemizethanks{\IEEEcompsocthanksitem{Jinxing Zhou, Dan Guo and Meng Wang are with Key Laboratory of Knowledge Engineering with Big Data (HFUT), Ministry of Education and School of Computer Science and Information Engineering, Hefei University of Technology, Hefei, China.}
\IEEEcompsocthanksitem{Xuyang Shen is with the Sensetime Research, Shanghai, China.}
\IEEEcompsocthanksitem{Jianyuan Wang is with Visual Geometry Group, University of Oxford, Oxford, United Kingdom.}
\IEEEcompsocthanksitem{Jiayi Zhang is with School of Computer Science and Engineering, the Beihang University, Beijing, China.}
\IEEEcompsocthanksitem{Weixuan Sun and Jing Zhang are with School of Computing, the Australian National University, Canberra, Australia.}
\IEEEcompsocthanksitem{Stan Birchfield is with Nvidia, Redmond, WA, USA.}
\IEEEcompsocthanksitem{Lingpeng Kong is with the University of Hong Kong, Hong Kong, China and Shanghai AI Lab, Shanghai, China.}
\IEEEcompsocthanksitem{Yiran Zhong is with Shanghai AI Lab, Shanghai, China.}
\IEEEcompsocthanksitem{\IEEEauthorrefmark{1}: These authors have equal contributions.}
\IEEEcompsocthanksitem{\IEEEauthorrefmark{2}: Meng Wang and Yiran Zhong are corresponding authors (e-mail: eric.mengwang@gmail.com, zhongyiran@gmail.com).}
}
}

\markboth{Journal of \LaTeX\ Class Files,~Vol.~14, No.~8, August~2015}%
{Shell \MakeLowercase{\textit{et al.}}: Bare Advanced Demo of IEEEtran.cls for IEEE Computer Society Journals}

\IEEEtitleabstractindextext{
\begin{abstract}
We propose a new problem called audio-visual segmentation (AVS), in which the goal is to output a pixel-level map of the object(s) that produce sound at the time of the image frame.
%
To facilitate this research, we construct the first audio-visual segmentation benchmark, \ie, AVSBench, providing pixel-wise annotations for sounding objects in audible videos. 
%
It contains three subsets: AVSBench-object (Single-source subset, Multi-sources subset) and AVSBench-semantic (Semantic-labels subset).
Accordingly, three settings are studied: 1) semi-supervised audio-visual segmentation with a single sound source; 2) fully-supervised audio-visual segmentation with multiple sound sources, and 3) fully-supervised audio-visual semantic segmentation.  
The first two settings need to generate binary masks of sounding objects indicating pixels corresponding to the audio, while the third setting further requires generating semantic maps indicating the object category. 
To deal with these problems, we propose a new baseline method that uses a temporal pixel-wise audio-visual interaction module to inject audio semantics as guidance for the visual segmentation process. 
We also design a regularization loss to encourage audio-visual mapping during training.
Quantitative and qualitative experiments on AVSBench compare our approach to several existing methods for related tasks, demonstrating that the proposed method is promising for building a bridge between the audio and pixel-wise visual semantics.
Code is available at {\color{blue}{\textit{\href{https://github.com/OpenNLPLab/AVSBench}{https://github.com/OpenNLPLab/AVSBench}}}}. Online benchmark is available at \color{blue}{\textit{\href{http://www.avlbench.opennlplab.cn}{http://www.avlbench.opennlplab.cn}}}. 

\end{abstract}

}

\maketitle

\IEEEdisplaynontitleabstractindextext
\IEEEpeerreviewmaketitle


%



\maketitle





\section{Introduction}
\label{intro}

\IEEEPARstart{H}{umans} largely rely on visual and auditory cues to understand their environmental surroundings. For example, a dog barking can be distinguished from a bird calling based on both their sound and appearance. Such audio-visual information is integrated with the brain in a synthesis process~\cite{wei2022learning}, crucial for comprehensively perceiving the world. Inspired by this cognitive ability of humans, we explore audio-visual learning with deep models via the integration of multi-modal signals.

\begin{figure*}[t]
\centering
\includegraphics[width=\textwidth]{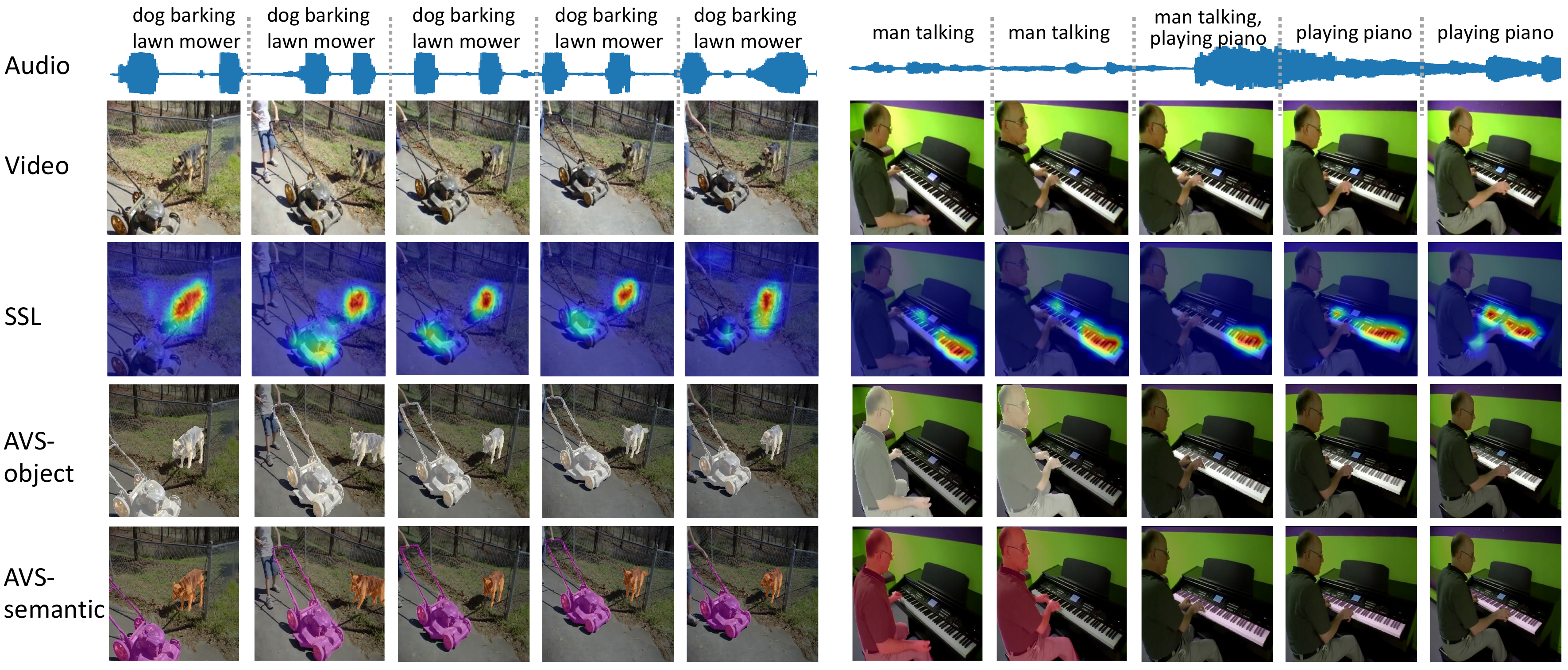} 
\vspace{-7mm}
\caption{\textbf{Comparison of the proposed AVS task with the Sound source localization (SSL) task.} SSL aims to estimate an approximate location of the sounding objects in the visual frame, at a patch level. In contrast, AVS estimates pixel-wise masks for all the sounding objects, regardless of the number of visible sounding objects. 
The segmentation masks can be binary or semantic under different task settings.
The binary masks indicate objects making sounds while the semantic masks further distinguish the object category. In the last row, the ground truths are displayed with the semantic masks.
}
\label{fig:task_comparison}
\vspace{-2mm}
\end{figure*}

Over the years, researchers have studied various problems within audio-visual artificial perception. For instance,
%
%
some researchers investigate the audio-visual correspondence (AVC) problem~\cite{arandjelovic2017look,arandjelovic2018objects,aytar2016soundnet}, which aims to determine whether an audio signal and a visual image describe the same scene. AVC is based on the phenomenon that these two signals usually occur simultaneously, such as a barking dog, a singing person, and a humming car. 
Others study the audio-visual event localization (AVEL)~\cite{lin2019dual,lin2020audiovisual,tian2018audio,wu2019dual,xu2020cross,ramaswamy2020makes,ramaswamy2020see,zhou2021positive,zhou2022cpsp,wang2022semantic}, which classifies the segments of a video using a set of pre-defined event labels.
Similarly, some research explores audio-visual video parsing (AVVP)~\cite{tian2020unified,wu2021exploring,lin2021exploring,yu2021mm,jiang2022dhhn,cheng2022joint,mo2022multi}, whose goal is to divide a video into several events and classify them as audible, visible, or both.
%
Due to a lack of pixel-level annotations, all these scenarios are restricted to the frame/temporal level, thus reducing the problem to audible image classification.

A related problem, known as sound source localization (SSL), aims to locate the visual regions within the image frames that correspond to the sound~\cite{arandjelovic2017look,arandjelovic2018objects,senocak2018learning,cheng2020look,owens2018audio,chen2021localizing,hu2019deep,qian2020multiple,hu2020discriminative}.
Compared to AVC/AVEL/AVVP, SSL seeks patch-level scene understanding, \ie, the results are usually presented by a heat map that is obtained either by visualizing the similarity matrix of the audio feature and the visual feature map, or by class activation mapping (CAM)~\cite{zhou2016cam}---without considering the actual shape of the sounding objects.

Building on this research, in this work we propose the pixel-level audio-visual segmentation (AVS) problem. 
This problem requires the network to densely predict whether each pixel corresponds to the given audio, so that a mask of the sounding object(s) is generated. Fig.~\ref{fig:task_comparison} illustrates the differences between SSL and AVS. 
As can be seen, the AVS task is more challenging as it requires the network to not only locate the audible frames but also delineate the shape of the sounding objects. 
Moreover, the AVS finally needs to classify the category semantics of different sounding objects. As shown in Fig.~\ref{fig:task_comparison}, each type of sounding object is assigned a specific color indicating its unique semantic category.

To facilitate this research, we release the AVSBench dataset, which is the first pixel-level audio-visual segmentation benchmark that provides ground truth labels for sounding objects. 
The dataset is divided into three subsets.
In the first subset, there is a single sound source in the video, leading to the task we call \textit{semi-supervised Single Sound Source Segmentation (S4)}.
In the second subset, there are multiple sound sources, leading to the task of \textit{fully-supervised Multiple Sound Source Segmentation (MS3)}.
For these two subsets, the ground truths are binary masks indicating pixels emitting the sounds. We study these two settings to have a basic perception of audio-visual segmentation from pixel-level.
While, the third subset is a Semantic-labels subset that introduces semantic labels of the sounding objects, exploring the task of \textit{fully-supervised Audio-Visual Semantic Segmentation (AVSS)}.
Compared to the S4 and MS3 settings, AVSS requires generating semantic maps that further tell the category information of the masked sounding objects.
As shown on the left example of Fig.~\ref{fig:task_comparison}, pixels of the \emph{dog} and \emph{lawn mower} are assigned with different colors indicating the unique semantic categories.
For convenience, we denote that the first two subsets, \ie, Single-source and Multi-sources subsets, constitute the AVSBench-object dataset, and the third Semantic-labels subset is also called the AVSBench-semantic dataset.
For all the settings, the goal is to segment the object(s) from the visual frames that are producing sounds.
Compared with traditional semantic segmentation~\cite{jon2014fcn,ron2015unet,zhong20183d,Wang2021PVTv2IB,xie2021segformer} task or video object segmentation~\cite{caelles2017one, perazzi2016benchmark, wang2020centermask} task, AVS is a multi-modal segmentation problem that necessitates the alignment of visual and audio semantics rather than classifying each pixel solely based on visual cues.

To deal with the aforementioned three settings, we test several methods from related tasks on AVSBench dataset and provide a new AVS method as a strong baseline. The framework is shown in Fig.~\ref{fig:framework}. It utilizes a standard encoder-decoder architecture but with a novel temporal pixel-wise audio-visual interaction (TPAVI) module to better introduce the audio semantics for guiding visual segmentation. We also propose a loss function to utilize the correlation of audio-visual signals, which further enhances segmentation performance.

At last, we remind that the audio-visual segmentation problem is first introduced in our previous work~\cite{zhou2022avs} that has been published in ECCV 2022.
Compared with the conference version, we add the following new extensions in this paper. 
\textbf{Firstly}, we expand upon our previous work by incorporating a new and challenging setting, \ie, the fully-supervised AVSS, which can be viewed as an independent task by the research community. 
%
We also conduct extensive ablation studies in this setting. 
These explorations help us gain a deeper understanding of audio-visual scenarios and design a more realistic model that enables us to perceive pixel-wise semantics.
\textbf{Secondly}, we propose the AVSBench-semantic dataset containing a Semantic-labels subset that newly provides pixel-wise semantic labels, as a significant complement of the original AVSBench dataset.
%
The AVSBench-semantic dataset includes significantly more event categories (70 \vs 23) and frames (80k \vs 10k) compared to the original AVSBench dataset.
%
More extending details of the video statistic and annotation are introduced in Sec.~\ref{sec:dataset}.
\textbf{Thirdly}, we update the previous AVS model~\cite{zhou2022avs} to predict the semantic maps and add extensive experiments on the new AVSBench dataset. For the convenience of the community, we build an online benchmark suite at \textcolor{blue}{\textit{\href{http://www.avlbench.opennlplab.cn}{http://www.avlbench.opennlplab.cn}}}.

\section{Related Work}\label{sec:related_work}

\textbf{Sound Source Localization (SSL).} Perhaps the most closely related problem to ours is SSL, which aims to locate the regions in the visual frames responsible for the sounds. The prediction of SSL is usually computed from the similarity matrix of the learned audio feature and the visual feature map~\cite{arandjelovic2017look,arandjelovic2018objects,senocak2018learning,cheng2020look,owens2018audio,chen2021localizing}, displayed as a heat map. 
SSL can also be divided into two settings according to the complexity of sound sources, \emph{viz.}, single and multiple sound source(s) localization. Here we focus on the challenging setting of multiple sources, which requires accurately localizing the true sound source among multiple potential candidates~\cite{hu2019deep,afouras2020self,qian2020multiple,hu2020discriminative}. 
In pioneering work, Hu \etal~\cite{hu2019deep} divide the audio and visual features into multiple cluster centers and take the center distance as a supervision signal to rank the paired audio-visual information.
Qian \etal ~\cite{qian2020multiple} first train an audio-visual correspondence model to extract coarse feature representations of audio and visual signals, and then use Grad-CAM~\cite{selvaraju2017grad} to visualize the class-specific features for localization. 
Furthermore, Hu \etal~\cite{hu2020discriminative} adopt a two-stage method, which first learns audio-visual semantics in the single sound source condition, using such learned knowledge to help with multiple sound sources localization.
Rouditchenko \etal ~\cite{rouditchenko2019self} tackles this problem by disentangling category concepts in the neural networks.
%
This method is actually more related to the task of \emph{sound source separation}~\cite{zhao2018sound,gao2018learning,zhao2019sound,gao2019co} and shows sub-optimal performance regarding visual localization.
Although these existing SSL methods indicate which regions in the image are making sound, the results do not clearly delineate the shape of the objects. 
Rather, the location map is computed by up-sampling the audio-visual similarity matrix from a low resolution.
Moreover, the methods above all rely on unsupervised learning when capturing the shape of sounding objects, which partly suffers from the lack of an annotated dataset.
To overcome these limitations, this paper provides an audio-visual segmentation dataset with pixel-level ground truth labels, which enables to achieve more accurate segmentation predictions.

{\noindent\textbf{Video Object Segmentation (VOS).}
The VOS task aims to segment the object of interest throughout the entire video sequence. It is divided into two settings: the semi-supervised and unsupervised. For the semi-supervised VOS, the target object is decided given a one-shot mask of the first sampled video frame~\cite{caelles2017one, perazzi2016benchmark, wang2020centermask}. As for unsupervised VOS, it needs to automatically segment all the primary objects~\cite{zhang2020unsupervised, faktor2014video, ventura2019rvos}.
Many excellent works are proposed and proven to achieve impressive segmentation performance~\cite{song2018pyramid, tokmakov2017learning, vijayanarasimhan2017sfm, chen2018blazingly, cheng2017learning, hu2017maskrnn}. However, these fancy designs are limited to a single visual modality. 
Recently, referring video object segmentation (R-VOS) attracts more attention~\cite{khoreva2018video, wu2022language, seo2020urvos, botach2022end}. The target object in R-VOS task is referred by a short language expression, whereas the proposed AVS task focuses on the audio-aligned visual objects, \ie, the object of interest is determined by the audio. Unlike the language used in R-VOS has clear semantics, the proposed AVSS requires a joint semantic classification for both audio and visual information, which makes it more challenging than the R-VOS task.
}

\noindent\textbf{Audio-Visual Dataset.}
To the best of our knowledge, there are no publicly available datasets that provide segmentation masks for the sounding visual objects with audio signals.
Here we briefly introduce the popular datasets in the audio-visual community.
For example, the AVE~\cite{tian2018audio} and LLP~\cite{tian2020unified} datasets are respectively collected for audio-visual event localization and video parsing tasks. 
They only have category annotations for video frames, and hence cannot be used for pixel-level segmentation. 
For the sound source localization problem, researchers usually use the Flickr-SoundNet~\cite{senocak2018learning} and VGG-SS~\cite{chen2021localizing} datasets, where the videos are sampled from the large-scale Flickr~\cite{aytar2016soundnet} and VGGSound~\cite{chen2020vggsound} datasets, respectively.
The authors provide bounding boxes to outline the location of the target sound source, which could serve as patch-level supervision. 
%
%
However, this still inevitably suffers from incorrect evaluation results since the sounding objects are usually irregular in shape and some regions within the bounding box actually do not correspond to the real sound source. 
The proposed AVSBench dataset provides pixel-wise semantic masks that accurately outline the shape of sounding objects.
This is beneficial for the research of pixel-level audio-visual learning.

\begin{table}[t]
\begin{center}
\caption{\textbf{AVSBench statistics}. The videos are split into train/valid/test.  The asterisk ($^*$) indicates one annotation per video whereas others are one annotation per second. $^\diamond$ in the last row indicates that 1,000 videos are withheld for online benchmarking.
}
\vspace{-3mm}
\label{table:dataset_split}
\setlength{\tabcolsep}{3.5pt}
\begin{tabular}{lcccc}
\toprule[0.8pt]\noalign{\smallskip}
subsets & classes & videos & train/valid/test  & \makecell[c]{labeled frames} \\
\noalign{\smallskip}
\midrule
\noalign{\smallskip}
single-source & 23 & \,4,932 & 3,452$^*$/740/740 & 10,852\\
multi-source & 23  & \,\,\,\,424   & 296/64/64 & \,\,\,2,120 \\
 semantic-labels & 70  & 12,356$^\diamond$   & 8,498/1,304/1,554 & 82,972 \\
\bottomrule[0.8pt]
\end{tabular}
\end{center}
\end{table}

\begin{table}[t]
\begin{center}
\caption{\textbf{Existing audio-visual dataset statistics.}
Each benchmark is shown with the number of videos and the \emph{annotated} frames. 
The final column indicates whether the frames are labeled by category, bounding boxes, or pixel-level masks. AVSBench extension provides pixel-level semantic labels with object category information.}
\label{table:comparison_with_datasets}
\setlength{\tabcolsep}{2pt}
\begin{tabular}{cccccc}
\toprule[0.8pt]\noalign{\smallskip}
benchmark & videos & frames & classes & types  &  annotations \\
\noalign{\smallskip}
\hline
AVE  \cite {tian2018audio}           & \,\,\,4,143 & 41,430  & \,\,\,28  & video  & category   \\ 
LLP \cite{tian2020unified}             & 11,849 & 11,849 & \,\,\,25  & video & category  \\
Flickr-SoundNet \cite{senocak2018learning} & \,\,\,5,000 & \,\,\,5,000  & \,\,\,50  & image & bbox   \\
VGG-SS \cite{chen2021localizing}         & \,\,\,5,158 & \,\,\,5,158  & 220  & image & bbox    \\  
AVSBench-object~\cite{zhou2022avs}       & \,\,\,5,356 & 12,972 & \,\,\,23   & video & pixel    \\
AVSBench-semantic       & 12,356 & 82,972 & \,\,\,70   & video & {pixel \&  category}    \\
\bottomrule[0.8pt]
\end{tabular}
\end{center}
\end{table}

\begin{figure*}[t]
\centering
\includegraphics[width=0.95\textwidth]{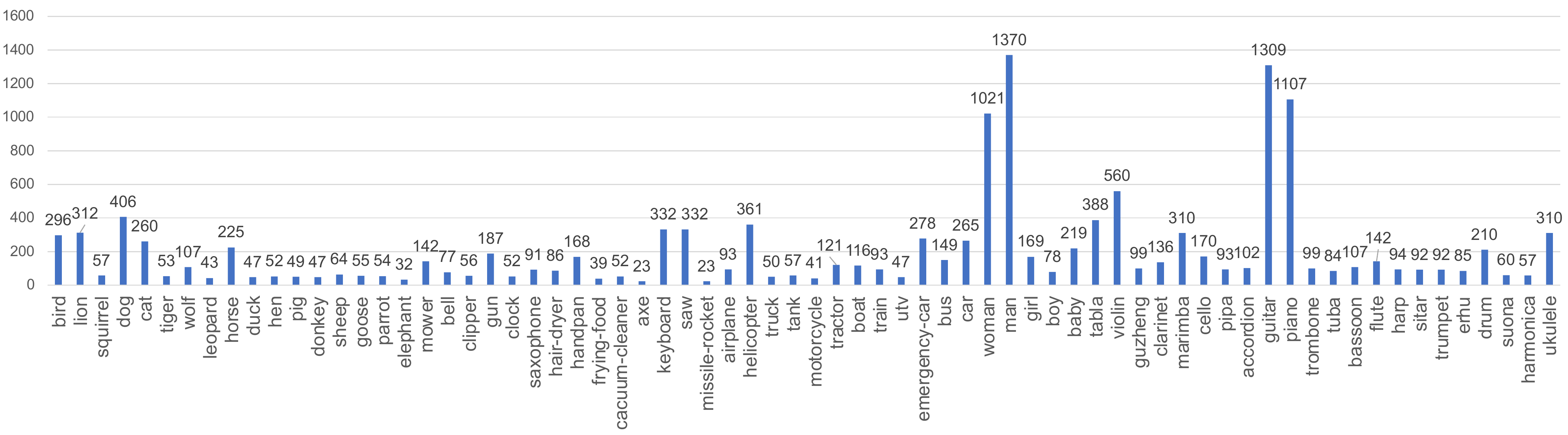}
\vspace{-3mm}
\caption{\textbf{Statistics of the AVSBench dataset extension, \ie, the AVSBench-semantic dataset.} There are 70 categories in the extension and the video number of each category is given.
}
\label{fig:single_source_statistics}
\vspace{-3mm}
\end{figure*}

\begin{figure*}[t]
\centering
\includegraphics[width=\textwidth]{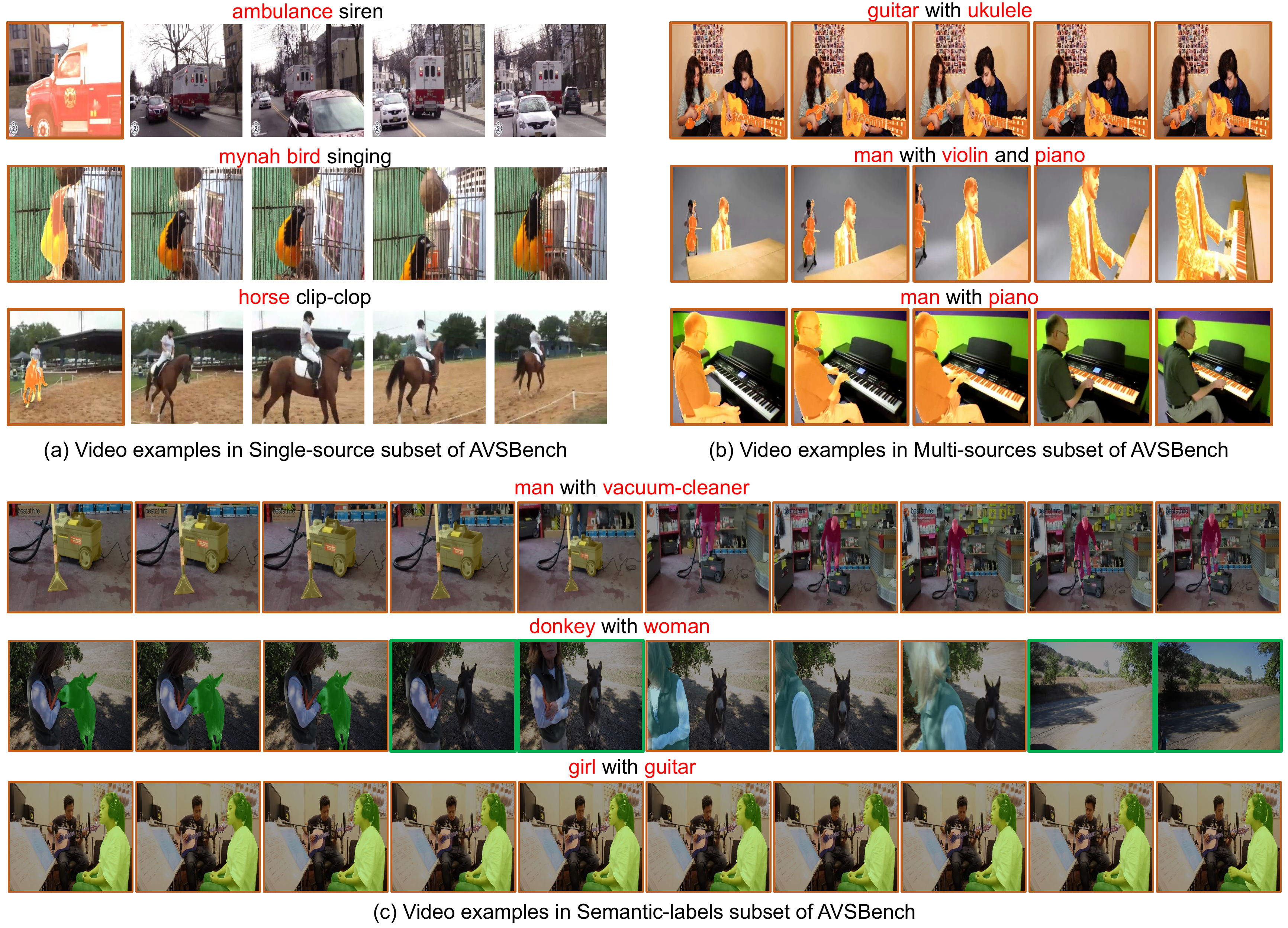}
\vspace{-6mm}
\caption{\textbf{AVSBench samples}. The AVSBench dataset contains the Single-source subset (a), Multi-sources subset (b), and Semantic-labels subset which mainly contains the multi-source videos (c).
Each video is divided into 5 clips for the first two, while 10 clips for the latter, as shown.
Annotated clips are indicated by brown framing rectangles while the green rectangles represent there are no sounding objects in those frames; the name of sounding objects is indicated by red text.
Binary masks of the sounding objects are annotated in the first two, reflected by the orange masks in (a) and (b).
The third subset provides colorful semantic masks indicating different object categories. 
Note that for the Single-source training set of AVSBench, only the first frame of each video is annotated, whereas all of the extracted frames are annotated for all other sets.
}
\label{fig:AVSbench_examples}
\vspace{-2mm}
\end{figure*}

\section{The AVSBench Dataset}\label{sec:dataset}
The AVSBench dataset is first proposed in our previous work ~\cite{zhou2022avs}. It contains a Single-source and a Multi-sources subset. Ground truths of these two subsets are binary segmentation maps indicating pixels of the sounding objects.
Recently, we collected a new Semantic-labels subset that provides semantic segmentation maps as labels.
We add it to the original AVSBench dataset as the third subset.
For convenience, we denote the original AVSBench dataset as \emph{AVSBench-object}, and the newly added Semantic-labels subset as \emph{AVSBench-semantic}.
In this section, we first introduce the video statistics and annotations of the AVSBench-object and then provide the extending details of AVSBench-semantic. 
Lastly, we introduce three benchmark settings on the updated AVSBench dataset.

\subsection{Dataset Statistics} 

\noindent\textbf{AVSBench-object.}
We collected the videos using the techniques introduced in VGGSound~\cite{chen2020vggsound} to ensure the audio and visual clips correspond to the intended semantics. AVSBench-object~\cite{zhou2022avs} contains two subsets---Single-source and Multi-sources---depending on the number of sounding objects. All videos were downloaded from YouTube under the \textit{Creative Commons} license, and each video was trimmed to $5$ seconds. The Single-source subset contains $4,932$ videos over $23$ categories, covering sounds from humans, animals, vehicles, and musical instruments. 
To collect the Multi-sources subset, we selected the videos that contain multiple sounding objects, \eg, a video of baby laughing, man speaking, and then woman singing. To be specific, we randomly chose two or three category names from the Single-source subset as keywords to search for online videos, then manually filtered out videos to ensure 1) each video has multiple sound sources, 2) the sounding objects are visible in the frames, and 3) there is no deceptive sound, \eg, canned laughter. In total, this process yielded $424$ videos for the Multi-sources subset out of more than six thousand candidates.
The ratio of train/validation/test split percentages is set as 70/15/15 for both subsets, as shown in Table~\ref{table:dataset_split}. 
Several video examples are visualized in Fig.~\ref{fig:AVSbench_examples}, where the red text indicates the name of sounding objects. 

\noindent\textbf{AVSBench-semantic.}
Since the original version of AVSBench dataset, \ie, AVSBench-object, we have extended the dataset by adding a third Semantic-labels subset that provides semantic segmentation maps as labels. AVSBench-semantic is enriched in video amount and audio-visual scene categories. In total, it contains 12,356 videos covering 70 categories. In Fig.~\ref{fig:single_source_statistics}, we show the category names and the video number for each category. This extension reserves all 5,356 videos from the original and upgrades them to 720p resolution. In addition, we further collect another 7,000 multi-source videos following the principle of collecting multi-sources subset of the original dataset. We reserve 1,000 videos for online evaluation and it will only be available for contestants in the future AVS Benchmark competition.
These newly collected videos are trimmed to 10 seconds which helps to train a segmentation model with the ability to encode long-range audio-visual sequences.
Except for the 1,000 withheld videos, the rest of the videos are split into 8,498 for training, 1,304 for validation, 1,554 for testing.
We also display some video examples in Fig.~\ref{fig:AVSbench_examples}.

The AVSBench-object and the AVSBench-semantic together form the updated AVSBench dataset.
We make a comparison between AVSBench with other popular audio-visual benchmarks in Table~\ref{table:comparison_with_datasets}.
The AVE~\cite{tian2018audio} dataset contains 4,143 videos covering 28 event categories. 
The LLP~\cite{tian2020unified} dataset consists of 11,849 YouTube video clips spanning $25$ categories, collected from AudioSet~\cite{gemmeke2017audio}.
Both the AVE and LLP datasets are labeled at a frame level, through audio-visual event boundaries.
Meanwhile, the Flickr-SoundNet~\cite{senocak2018learning} dataset and VGG-SS~\cite{chen2021localizing} dataset are proposed for sound source localization (SSL), labeled at a patch level through bounding boxes.
{{The AVSBench-object (original AVSBench dataset~\cite{zhou2022avs}) contains 5,356 videos with 12,972 pixel-wise annotated frames which is designed to facilitate research on fine-grained audio-visual segmentation.
AVSBench-semantic further extends it from three aspects: 1) the video quantity is expanded to 12,356 and focuses more on the multi-source case; 2) the number of object categories is enlarged from 23 to 70; 3) annotations are updated from pixel-wise binary mask to semantic masks.}} 
The recent AVSBench dataset provides accurate semantic maps as ground truth. This makes it beneficial not only for the proposed audio-visual segmentation but also for sound source localization, which could help the training of SSL methods and serve as an evaluation benchmark.


\subsection{Annotation}

\noindent\textbf{AVSBench-object.}
Videos in AVSBench-object are trimmed to 5 seconds. We divide each 5-second video into five equal 1-second clips, and we provide manual pixel-level annotations for the 1-second clips. The ground truth label is a binary mask indicating the pixels of sounding objects, according to the audio at the corresponding time. For example, in the Multi-sources subset, even though a dancing person shows drastic movement spatially, it would not be labeled as long as no sound was made. 
In clips where objects do not make sound, the object should not be masked, \eg, the \emph{piano} in the first two clips of the last row of Fig.~\ref{fig:AVSbench_examples}b. Similarly, when more than one object emits sound, all the emitting objects are annotated, \eg, the guitar and ukulele in the
first row in Fig.~\ref{fig:AVSbench_examples}b.
Also, when the sounding objects in the video are changing dynamically, the difficulty is further increased, \eg, the second, third, and fourth rows in Fig.~\ref{fig:AVSbench_examples}b.

We use two types of labeling strategies, based on the different difficulties between the Single-source and the Multi-sources subsets. For the videos in the training split of Single-source, we only annotate the first sampled frame (with the assumption that the information from one-shot annotation is sufficient, as the Single-source subset has a single and consistent sounding object over time). This assumption is verified by the quantitative experimental results shown in Table~\ref{table:comparison_with_baselines}.
For the more challenging Multi-sources subset, all clips are annotated for training, since the sounding objects may change over time. 
Note that for validation and test splits, all clips are annotated, as shown in Table~\ref{table:dataset_split}.

\noindent\textbf{AVSBench-semantic.}
The AVSBench-semantic subset uses the videos from AVSBench-object. For these videos, we update the annotated binary masks to semantic masks by adding category information of the sounding objects. As for the newly collected 10-second videos, we sample ten video frames and provide their semantic annotations, similar to the annotation process of AVSBench-object. We show some annotation examples in Fig.~\ref{fig:AVSbench_examples}(c). As shown, the sounding object is highlighted with unique color indicating its category. Also, when there is no sound or the sounding object is out of the screen (green boxes in the second row of Fig.~\ref{fig:AVSbench_examples}(c)), that video frame will not be annotated.

\subsection{Benchmark Setting}\label{sec:problem_statement}
{{We provide three benchmark settings:
the semi-supervised Single Sound Source Segmentation (S4), the fully supervised Multiple Sound Source Segmentation (MS3), and the fully supervised audio-visual semantic segmentation (AVSS). The former two settings are based on the AVSBench-object dataset while the AVSS is conducted on the AVSBench-semantic.}} For ease of expression, we denote the video sequence as $S$, which consists of $T$ non-overlapping yet continuous clips $\{S_t^v, S_t^a\}_{t=1}^{T}$, where $S^v$ and $S^a$ are the visual and audio components, $T$ is equal to 5 for AVSBench-object while 10 for AVSBench-semantic. In practice, we extract the video frame at the end of each second.

\noindent\textbf{Semi-supervised S4} corresponds to the Single-source subset of AVSBench-object. It is termed as semi-supervised because only part of the ground truth is given during training (\ie, the first sampled frame of the videos) but all the video frames require a prediction during evaluation. We denote the pixel-wise label as $\bm{Y}_{t=1}^s \in \mathbb{R}^{H \times W}$, where $H$ and $W$ are the frame height and width, respectively. $\bm{Y}_{t=1}^s$ is a binary matrix where $1$ indicates sounding objects while $0$ corresponds to background or silent objects. 

\noindent\textbf{Fully-supervised MS3} deals with the Multi-sources subset of AVSBench-object, where the labels of all five frames of each video are available for training. The ground truth is denoted as $\{\bm{Y}_t^m\}_{t=1}^{T}$, where $\bm{Y}_t^{m} \in \mathbb{R}^{H \times W}$ is the binary label for the $t$-th video clip.

\begin{figure*}[t]
\centering
\includegraphics[width=\textwidth]{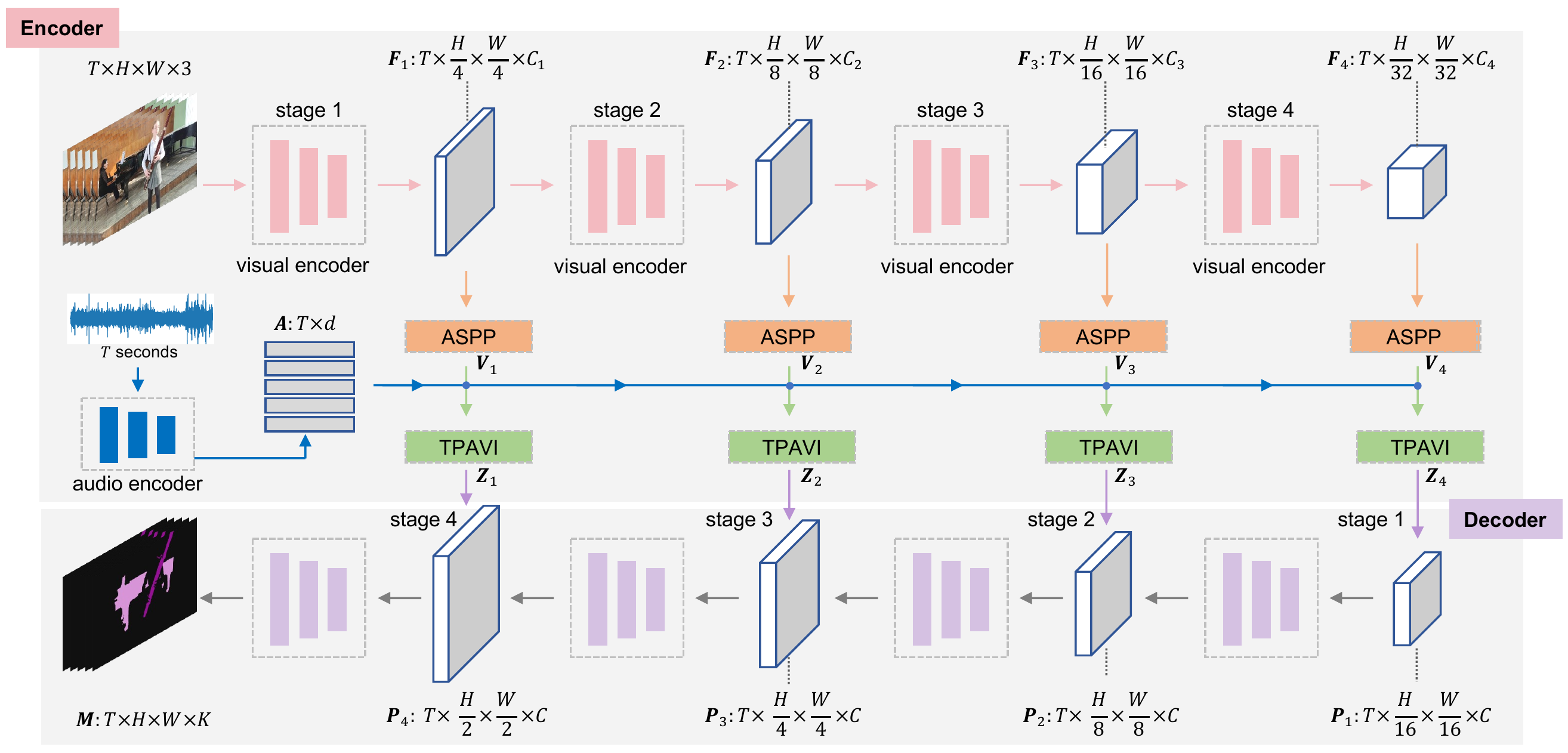}
\vspace{-6mm}
\caption{\textbf{Overview of the Baseline}, which follows a hierarchical {Encoder-Decoder} pipeline. The \emph{encoder} takes the video frames and the entire audio clip as inputs, and outputs visual and audio features, respectively denoted as $\bm{F}_i$ and $\bm{A}$.  The visual feature map $\bm{F}_i$ at each stage is further sent to the ASPP~\cite{chen2017deeplab} module and then our TPAVI module (introduced in Sec.~\ref{sec:approach}). ASPP provides different receptive fields for recognizing visual objects, while TPAVI focuses on the temporal pixel-wise audio-visual interaction. The \emph{decoder} progressively enlarges the fused feature maps by four stages and finally generates the output mask $\bm{M}$ for sounding objects.}
\label{fig:framework}
\end{figure*}

{{\noindent\textbf{{Fully-supervised} AVSS} deals with the Semantic-labels subset of AVSBench-semantic, where the semantic masks of all ten frames of each video are known during training. The ground truth can be denoted as $\{\bm{Y}_t\}_{t=1}^{T}$, where $\bm{Y}_t \in \mathbb{R}^{H \times W \times K}$ is the semantic label for the $t$-th video clip, $K$ is the total category number of the sounding objects in the dataset. }}

The goal for all the settings is to correctly segment the sounding object(s) for each video clip by utilizing the audio and visual cues, \ie, $S^a$ and $S^v$. {Different from S4 and MS3 settings, AVSS setting needs to further output the category of the sounding objects.} Generally, it is expected $S^a$ to indicate the target object, while $S^v$ provides information for fine-grained segmentation. 
The predictions are denoted as $\{\bm{M}_t\}_{t=1}^{T}$, $\bm{M}_t \in \mathbb{R}^{H \times W \times K}$, where $K=1$ under S4 and MS3 settings. 

\section{A Baseline}\label{sec:approach}
We propose a new baseline method for the pixel-level audio-visual segmentation problem as shown in Fig.~\ref{fig:framework}.
Following the convention of semantic segmentation methods \cite{jon2014fcn,ron2015unet,Wang2021PVTv2IB,xie2021segformer}, our method adopts an encoder--decoder architecture. 




\noindent\textbf{The Encoder:} We extract audio and visual features independently. Given an audio clip $S^a$, we first process it to a spectrogram via the short-time Fourier transform and then send it to a convolutional neural network, VGGish~\cite{hershey2017cnn}. We use the weights that are pretrained on AudioSet~\cite{gemmeke2017audio} to extract audio features $\bm{A} \in \mathbb{R}^{T \times d}$, where $d=128$ is the feature dimension.
For a video frame $S^v$, we extract visual features with popular convolution-based or vision transformer-based backbones. We try both two options in the experiments and they show similar performance trends. These backbones produce hierarchical visual feature maps during the encoding process, as shown in Fig.~\ref{fig:framework}. 
We denote the features as $\bm{F}_i \in \mathbb{R}^{T \times h_i \times w_i \times C_i}$, where $(h_i,w_i)=(H,W)/2^{i+1}$, $i=1,\ldots,n$. 
The number of levels is set to $n=4$ in all experiments.


\noindent\textbf{Cross-Modal Fusion:} We use Atrous Spatial Pyramid Pooling (ASPP) modules~\cite{chen2017deeplab} to further post-process the visual features $\bm{F}_i$ to $\bm{V}_i \in \mathbb{R}^{T \times h_i \times w_i \times C}$, where $C=256$. These modules employ multiple parallel filters with different rates and hence help to recognize visual objects with different receptive fields, \eg, different-sized moving objects.

\begin{figure}
\begin{center}
\includegraphics[width=0.4\textwidth]{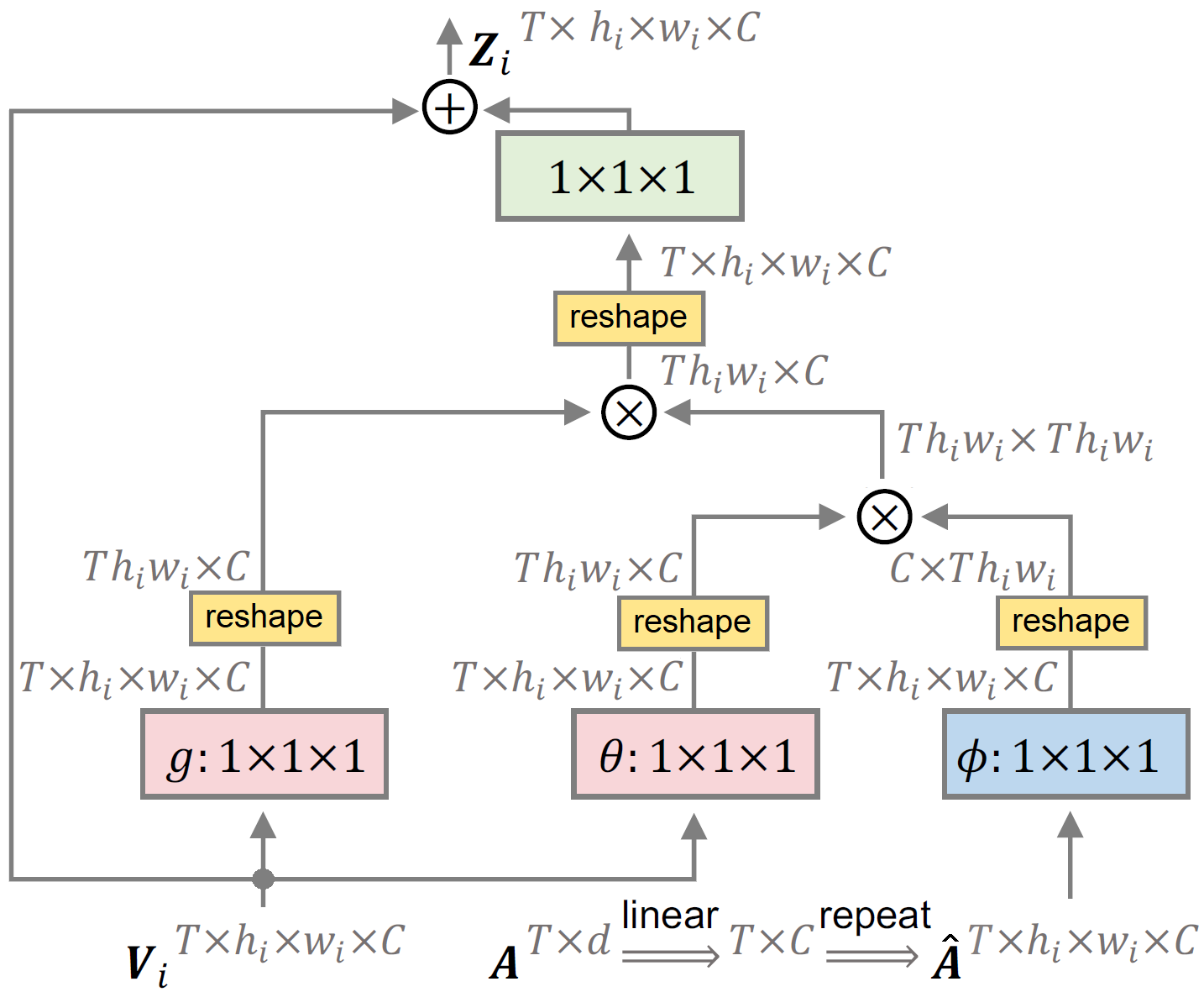}
  \caption{\textbf{The TPAVI module} takes the $i$-th stage visual feature $\bm{V}_i$ and the audio feature $\bm{A}$ as inputs. The colored boxes represent $1 \times 1 \times 1$ convolutions, while the yellow boxes indicate reshaping operations. The symbols ``$\otimes$'' and ``$\oplus$'' denote matrix multiplication and element-wise addition, respectively.}\label{fig:TPAVI_module}
  \end{center}
\end{figure}

Then, we consider introducing the audio information to build the audio-visual mapping to assist with identifying the sounding object. This is particularly essential for the MS3 and AVSS settings where there are multiple dynamic sound sources.
Our intuition is that, although the auditory and visual signals of the sound sources may not appear simultaneously, they usually exist in more than one video frame.
Therefore, integrating the audio and visual signals of the whole video should be beneficial. 
%
%
Motivated by~\cite{wang2018non} that uses the non-local block to encode space-time relation, we adopt a similar module to encode the temporal pixel-wise audio-visual interaction (TPAVI).
As illustrated in Fig.~\ref{fig:TPAVI_module}, the current visual feature map $\bm{V}_i$ and the audio feature $\bm{A}$ of the entire video are sent into the TPAVI module.
Specifically, the audio feature $\bm{A}$ is first transformed to a feature space with the same dimension as the visual feature $\bm{V}_i$, by a linear layer.
Then it is spatially duplicated $h_i w_i$ times and reshaped to the same size as $\bm{V}_i$. We denote such processed audio feature as $\hat{\bm{A}}$. Next, it is expected to find those pixels of visual feature map $\bm{V}_i$ that have a high response to the audio counterpart $\hat{\bm{A}}$ through the entire video. 

Such an audio-visual interaction can be measured by dot-product, then the updated feature maps $\bm{Z}_i$ at the $i$-th stage can be computed as,
\begin{equation}\label{eq:tpavi}
\footnotesize
\textstyle
\begin{aligned}
    & \bm{Z}_i = \bm{V}_i + \mu(\alpha_i\ {g(\bm{V}_i})), \ \text{where} \ \alpha_i = \frac{\theta(\bm{V}_i) \ {\phi(\hat{\bm{A}})}^{\top}}{N} \\
\end{aligned}
\end{equation}
\noindent where $\theta$, $\phi$, $g$ and $\mu$ are  $1 \times 1 \times 1$ convolutions, $N=T \times h_i \times w_i$ is a normalization factor, $\alpha_i$ denotes the audio-visual similarity, and $\bm{Z}_i \in \mathbb{R}^{T \times h_i \times w_i \times C}$. Each visual pixel interacts with all the audio through the TPAVI module. 
 We provide a visualization of the audio-visual attention in TPAVI later in Fig.~\ref{fig:audio_visual_attention}, which shows a similar ``appearance'' to the prediction of SSL methods because it constructs a pixel-to-audio mapping.


\noindent\textbf{The Decoder:} We adopt the decoder of Panoptic-FPN~\cite{kirillov2019panoptic} in this work for its flexibility and effectiveness, though any valid decoder architecture could be used. In short, at the $j$-th stage, where $j=2,3,4$, both the outputs from stage $\bm{Z}_{5-j}$ and the last stage $\bm{Z}_{6-j}$ of the encoder are utilized for the decoding process. The decoded features are then upsampled to the next stage. {The final output of the decoder is $\bm{M} \in \mathbb{R}^{T \times H \times W \times K}$. For S4 and MS3 settings, $K=1$, and the output is then activated by \emph{sigmoid} function. 
During inference of the AVSS setting, the output is further processed by a \emph{softmax} operation along the $K$ channel, and the index with the highest probability represents the category of the sounding object.}


\begin{table*}[t]
\caption{\textbf{Comparison with methods from related tasks on audio-visual segmentation under the S4 and MS3 settings.}
The compared methods come from the tasks of sound source localization (SSL), video object segmentation (VOS), and salient object detection (SOD). Results of mIoU~(\%) and F-score are reported.}
\begin{center}
 \begin{threeparttable}
  \begin{tabular}{lp{1.4cm}<{\centering}p{1.4cm}<{\centering}p{1.4cm}<{\centering}p{1.4cm}<{\centering}p{1.4cm}<{\centering}p{1.4cm}<{\centering}p{1.4cm}<{\centering}p{1.4cm}<{\centering}p{1.4cm}<{\centering}p{1.4cm}<{\centering}}
  \toprule[0.8pt]
      \multirow{2}{*}{Metric} & \multirow{2}{*}{Setting}   & \multicolumn{2}{c}{SSL}  & \multicolumn{2}{c}{VOS} & \multicolumn{2}{c}{SOD} & \multicolumn{2}{c}{AVS}\\ 
    \cmidrule(r){3-4}\cmidrule(r){5-6}\cmidrule(r){7-8}\cmidrule(r){9-10}
                   & & LVS\cite{chen2021localizing} & MSSL\cite{qian2020multiple} & 3DC\cite{mahadevan2020making} & SST\cite{duke2021sstvos} & iGAN\cite{mao2021transformer} & LGVT\cite{zhang2021learning} & ResNet50 & PVT-v2 \\ \midrule
     \multirow{2}{*}{mIoU} & S4 & 37.94 & 44.89 & 57.10 & 66.29 & 61.59 & 74.94 & 72.79 & \textbf{78.74}  \\
    & MS3  & 29.45 & 26.13 & 36.92 & 42.57  & 42.89 & 40.71 & 47.88 & \textbf{54.00}  \\ \midrule
     \multirow{2}{*}{F-score} & S4 & .510 & .663 & .759 & .801 & .778 & .873 & .848 & \textbf{.879}  \\
    & MS3  & .330 & .363 & .503  &   .572& .544 & .593 &  .578 & \textbf{.645}  \\
    \bottomrule[0.8pt]
  \end{tabular}
\end{threeparttable}
\end{center}\label{table:comparison_with_baselines}
\end{table*}

\noindent\textbf{Objective function:}\label{sec:loss_function}
Given the prediction $\bm{M}$ and the pixel-wise label $\bm{Y}$, we adapt the binary cross entropy (BCE) loss as the main supervision function. Besides, we use an additional regularization term $\mathcal{L}_\text{AVM}$ to force the audio-visual mapping. Specifically, we use the Kullback–Leibler (KL) divergence to ensure the masked visual features have similar distributions with the corresponding audio features. In other words, if the audio features of some frames are close in the feature space, the corresponding sounding objects are expected to be close in the feature space. The total objective function $\mathcal{L}$ can be computed as follows:
\begin{align}\label{eq:loss}
  &\mathcal{L} = \text{BCE}(\bm{M}, \bm{Y}) + \lambda \mathcal{L}_\text{AVM}(\bm{M},\bm{Z},\bm{A}), \\
  &\mathcal{L}_\text{AVM} = \sum_{i=1}^{n}( \text{KL}( avg \ ( \bm{M}_i \odot \bm{Z}_i), \bm{A}_i), \label{eq:avmloss}
\end{align}
where $\lambda$ is a balance weight, $\odot$ denotes element-wise multiplication, and $\textit{avg}$ denotes the average pooling operation. 
At each stage, we down-sample the prediction $\bm{M}$ to $\bm{M}_i$ via average pooling to have the same shape as $\bm{Z}_i$.
The vector $\bm{A}_i$ is a linear transformation of $\bm{A}$ that has the same feature dimension with $\bm{Z}_i$.
%
For the semi-supervised S4 setting, we found that the audio-visual regularization loss does not help, so we set $\lambda=0$ in this setting.



\section{Experimental Results}
\subsection{Implementation details}\label{sec:experimental_setup}


We conduct training and evaluation on the upgraded AVSBench dataset, with both convolution-based and transformer-based backbones, ResNet-50~\cite{he2016deep} and Pyramid Vision Transformer~(PVT-v2)~\cite{Wang2021PVTv2IB}.
Both of the backbones are pretrained on the ImageNet~\cite{russakovsky2015imagenet} dataset.
All the video frames are resized to a shape of $224 \times 224$. 
%
%
The channel sizes of the four stages are $C_{1:4} = [256, 512, 1024, 2048]$ and $C_{1:4} = [64, 128, 320, 512]$ for ResNet-50 and PVT-v2, respectively. 
The channel size of the ASPP module is set to $C=256$.
%
We use the VGGish model to extract audio features, a VGG-like network~\cite{hershey2017cnn} pretrained on the AudioSet~\cite{gemmeke2017audio} dataset.
The audio signals are converted to one-second splits as the network inputs.
%
%
We use the Adam optimizer with a learning rate of 1e-4 for training.
The batch size is set to $4$ and the number of training epochs are $15$, $30$, and $60$ respectively for the semi-supervised S4, the fully-supervised MS3, and the AVSS settings. 
The $\lambda$ in Eq.~\eqref{eq:loss} is empirically set to 0.5. 
%

\subsection{Comparison with methods from related tasks}\label{sec:exp_compare_to_ssl}

Predictions under the S4 and MS3 settings are binary segmentation maps while they are semantic maps under the AVSS setting.
Methods from different related tasks need to be compared under these settings.
We introduce the comparison results of the former two settings in Sec.~\ref{sec:exp_s4_ms3} and the AVSS setting in Sec.~\ref{sec:exp_avss}.

\subsubsection{Comparison under the S4 and MS3 settings}\label{sec:exp_s4_ms3}
{For the audio-visual segmentation under S4 and MS3 settings}, we compare our baseline framework with the methods from three related tasks, including sound source localization (SSL), video object segmentation (VOS), and salient object detection (SOD). 
For each task, we report the results of two SOTA methods on our AVSBench-object dataset, \ie, LVS~\cite{chen2021localizing} and MSSL~\cite{qian2020multiple} for SSL, 3DC~\cite{mahadevan2020making} and SST~\cite{duke2021sstvos} for VOS, iGAN~\cite{mao2021transformer} and LGVT~\cite{zhang2021learning} for SOD. We select these methods as they are state-of-the-art in their fields: 1) \textit{LVS} uses the background and the most confident regions of sounding objects to design a contrastive loss for audio-visual representation learning and the localization map is obtained by computing the audio-visual similarity.
2) \textit{MSSL} is a two-stage method for multiple sound source localization and the localization map is obtained by Grad-CAM~\cite{selvaraju2017grad}.
3) \textit{3DC} adopts an architecture that is fully constructed by powerful 3D convolutions to encode video frames and predict segmentation masks.
4) \textit{SST} introduces a transformer architecture to achieve sparse attention of the features in the spatiotemporal domain.
5) \textit{iGAN} is a ResNet-based generative model for saliency detection, considering about the inherent uncertainty of saliency detection.
6) \textit{LGVT} is a saliency detection method based on Swin transformer \cite{liu2021swin}, whose long-range dependency modeling ability  leads to better global context modeling.
We adopt the architecture of these methods and fit them into our semi-supervised S4 and fully-supervised MS3 settings. 
For a fair comparison, the backbones of these methods are all pretrained on the ImageNet~\cite{russakovsky2015imagenet}. 

\begin{figure*}[t]
\centering
\includegraphics[width=\textwidth]{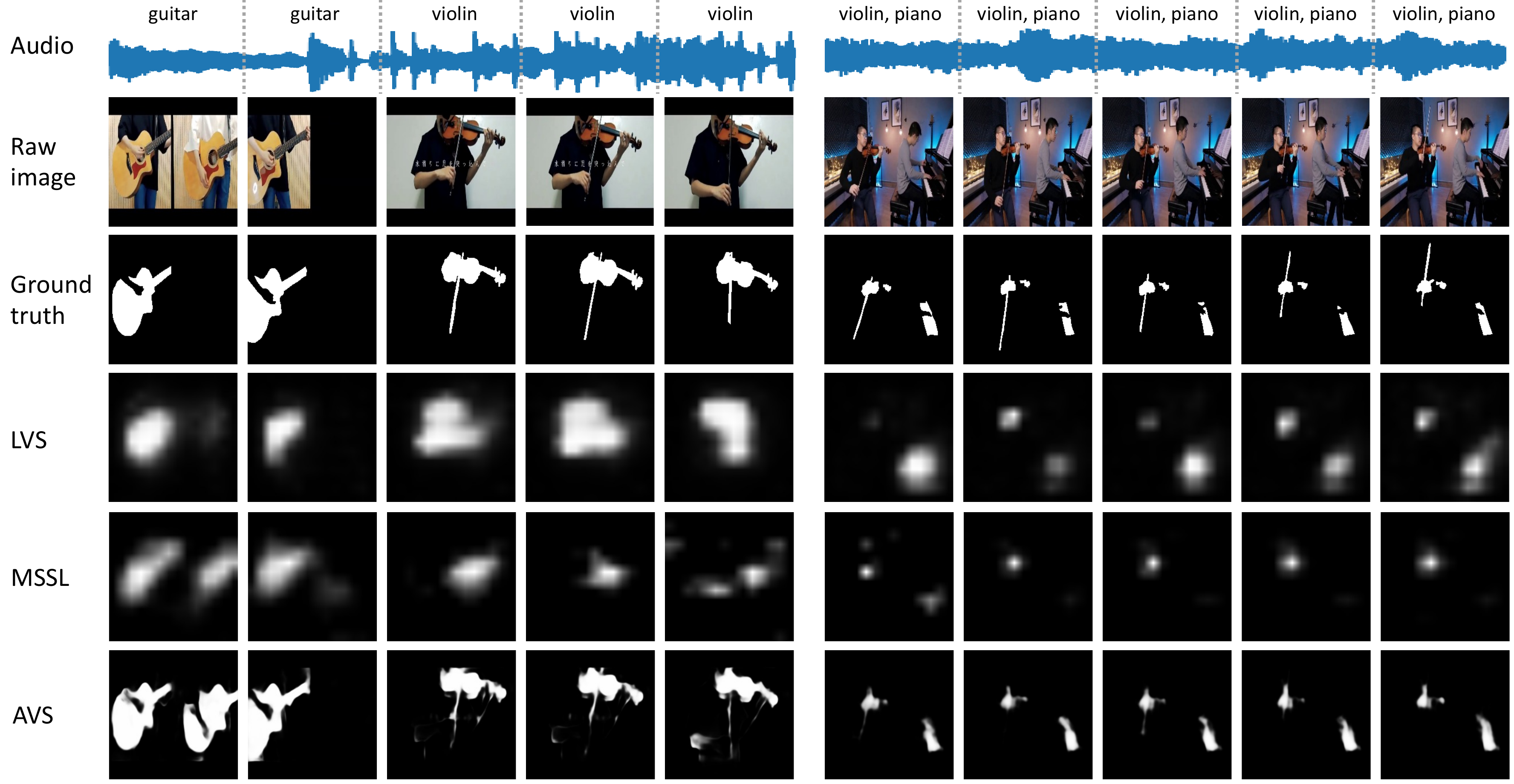}
\vspace{-6mm}
\caption{\textbf{Qualitative examples of the SSL methods and our AVS framework under the fully-supervised MS3 setting.} The SSL methods (LVS \cite{chen2021localizing} and MSSL \cite{qian2020multiple}) can only generate rough location maps, while the AVS framework can accurately segment the pixels of sounding objects and nicely outline their shapes.}
\label{fig:example_avs_ssl}
\end{figure*}

\noindent\textbf{Quantitative comparison between AVS and SSL/SOD/VOS.}
The quantitative results are shown in Table~\ref{table:comparison_with_baselines}, with Mean Intersection over Union (mIoU) and F-score\footnote{F-score considers both the precision and recall: $F_\beta = \frac{(1+\beta^2) \times \mathsf{precision} \times \mathsf{recall}}{\beta^2 \times \mathsf{precision} + \mathsf{recall}}$, where $\beta^2$ is set to 0.3 in our experiments.}.
%
There is a substantial gap between the results of SSL methods and those of our baseline, mainly because the SSL methods cannot provide a pixel-level prediction.
Also, our baseline framework shows a consistent superiority to the VOS and SOD methods in both semi-supervised S4 and fully-supervised MS3 settings.
%
It is worth noting that the state-of-the-art SOD method LGVT~\cite{zhang2021learning} slightly outperforms our ResNet50-based baseline under the Single-source set (74.94\% mIoU \textit{vs.} 72.79\% mIoU), mainly because LGVT uses the strong Swin Transformer backbone~\cite{liu2021swin}.
However, when it comes to the Multi-sources setting, the performance of LGVT is obviously worse than that of our ResNet50-based baseline (40.71\% mIoU \textit{vs.} 47.88\% mIoU).
This is because the SOD method relies on the dataset prior, and cannot handle situations where sounding objects change but visual contents remain the same.
Instead, the audio signals guide our method to identify which object to segment, leading to better performance.
Moreover, if also using a transformer-based backbone, our method is stronger than LGVT in both settings.
Besides, we notice that although SSL methods utilize both audio and visual signals, they cannot match the performance of VOS or SOD methods that only use visual frames.
It indicates the significance of pixel-wise scene understanding. 
%
The proposed AVS baselines achieve satisfactory performance under the semi-supervised S4 setting (around 70\% mIoU), which verifies that one-shot annotation is sufficient for single-source cases.
%


\begin{figure*}[t]
\centering
\includegraphics[width=\textwidth]{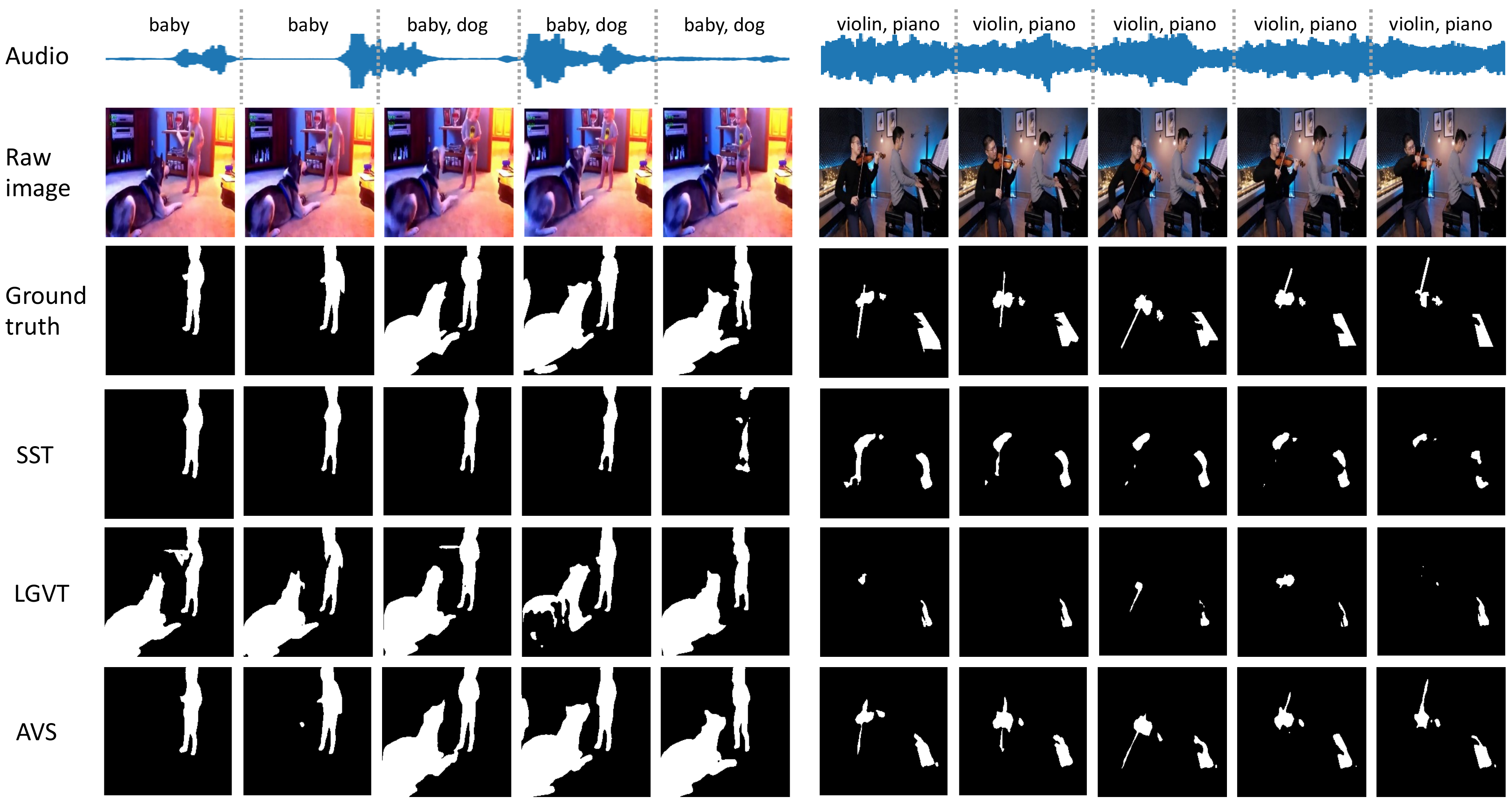} 
\vspace{-6mm}
\caption{\textbf{Qualitative examples of the VOS, SOD, and our AVS methods under the fully-supervised MS3 setting.} We pick the state-of-the-art VOS method SST~\cite{duke2021sstvos} and SOD method LGVT~\cite{zhang2021learning}. As can be verified in the left sample, SST or LGVT cannot capture the change of sounding objects (from `baby' to `baby and dog'), while the AVSS accurately conducts prediction under the guidance of the audio signal.
%
}
\label{fig:example_avs_sod}
\vspace{-3mm}
\end{figure*}

\noindent\textbf{Qualitative comparison between AVS and SSL/VOS/SOD.}
We provide some qualitative examples to compare our AVS framework with the SSL methods, LVS~\cite{chen2021localizing} and MSSL~\cite{qian2020multiple}. As shown in the left sample of Fig.~\ref{fig:example_avs_ssl},
LVS over-locates the sounding object \emph{violin}.
At the same time, MSSL fails to locate the \emph{piano} of the right sample.
Both the results of these two methods are blurry and they cannot accurately locate the sounding objects.
Instead, the proposed AVS framework can not only accurately segment all the sounding objects, but also nicely outline the object shapes.

Besides, we also compare the proposed AVS framework with the state-of-the-art methods from VOS and SOD, \ie, SST~\cite{duke2021sstvos} and LGVT~\cite{zhang2021learning}, respectively.
As shown in Fig.~\ref{fig:example_avs_sod},  SST and LGVT can predict their objects of interest in a pixel-wise manner.
However, their predictions rely on the visual saliency and the dataset prior, which cannot satisfy our problem setting. 
For example, in the left sample of Fig.~\ref{fig:example_avs_sod}, the \emph{dog} keeps quiet in the first two frames and should not be viewed as an object of interest in our problem setting.
Our AVS method correctly follows the guidance of the audio signal, \ie, accurately segmenting the \emph{baby} at the first two frames and both the sounding objects at the last three frames, with their shapes complete.
Instead, the VOS method SST misses the barking dog at the last three frames.
The SOD method LGVT masks out both the \emph{baby} and \emph{dog} over all the frames mainly because these two objects usually tend to be `salient', which is not desired in this sample.
When it comes to the right sample of Fig.~\ref{fig:example_avs_sod}, we can observe that LGVT almost fails to capture the \emph{violin}, since the violin is relatively small.
The VOS method SST can find the rough location of the violin, with the help of the information from temporal movement.
In contrast, our AVS framework can accurately depict the shapes and locations of the violin and piano.

\subsubsection{Comparison under the AVSS setting}\label{sec:exp_avss}
For the audio-visual semantic segmentation (AVSS) setting, the experiments are conducted on the Semantic-labels subset.
We compare the proposed baseline to two methods from the VOS task since they can also generate semantic maps from videos.
Specifically, we include the aforementioned 3DC~\cite{mahadevan2020making} and a newly proposed SOTA method AOT~\cite{yang2021aot} in our comparison. We select AOT as a referenced method because it proposes a new long-short term transformer layer and can effectively handle multi-object scenarios, whereas our AVS model also focuses on multiple sounding objects.

\noindent\textbf{Quantitative comparison between AVS and VOS.}
As shown in Table~\ref{table:avss_vos}, the strong AOT model surpasses our ResNet50-based AVS model but the PVT-based AVS model keeps the top performance (29.77\% mIoU, 0.352 F-score).
Besides, we found the performance under the AVSS setting is much lower than the S4 and MS3 settings.
For example, the mIoU is 54.00\% under the MS3 setting while it is 29.77\% under the AVSS setting, using the same PVT-v2 based backbone.
One of the main reason should be that the AVSS setting needs to further predict the category semantic of each pixel. Notably, there are more multi-source videos covering 70 classes in the dataset and some objects are hard to identify in appearance or sound.

\begin{table}[t]
\small
\caption{\textbf{Comparison with methods from VOS task on audio-visual segmentation under the AVSS setting.} Results of mIoU (\%) and F-score are reported.}
\begin{center}
 \begin{threeparttable}
  \begin{tabular}{   p{1.3cm}p{1.3cm}<{\centering}p{1.3cm}<{\centering}p{1.3cm}<{\centering}p{1.5cm}<{\centering}}
  \toprule[0.8pt]
      \multirow{2}{*}{Metric} &  \multicolumn{2}{c}{VOS}  & \multicolumn{2}{c}{AVS} \\ 
    \cmidrule(r){2-3}\cmidrule(r){4-5} 
                   & {3DC~\cite{mahadevan2020making}} & {AOT~\cite{yang2021aot}} & {ResNet50} & {PVT-v2} \\ \midrule
     {mIoU} &  17.27 & 25.40 & 20.18 & \textbf{29.77}  \\
    {F-score}  & .216  & .310 &  .252 & \textbf{.352}  \\ 
    \bottomrule[0.8pt]
  \end{tabular}
\end{threeparttable}
\end{center}\label{table:avss_vos}
\end{table}

\begin{figure*}[t]
\centering
\includegraphics[width=\textwidth]{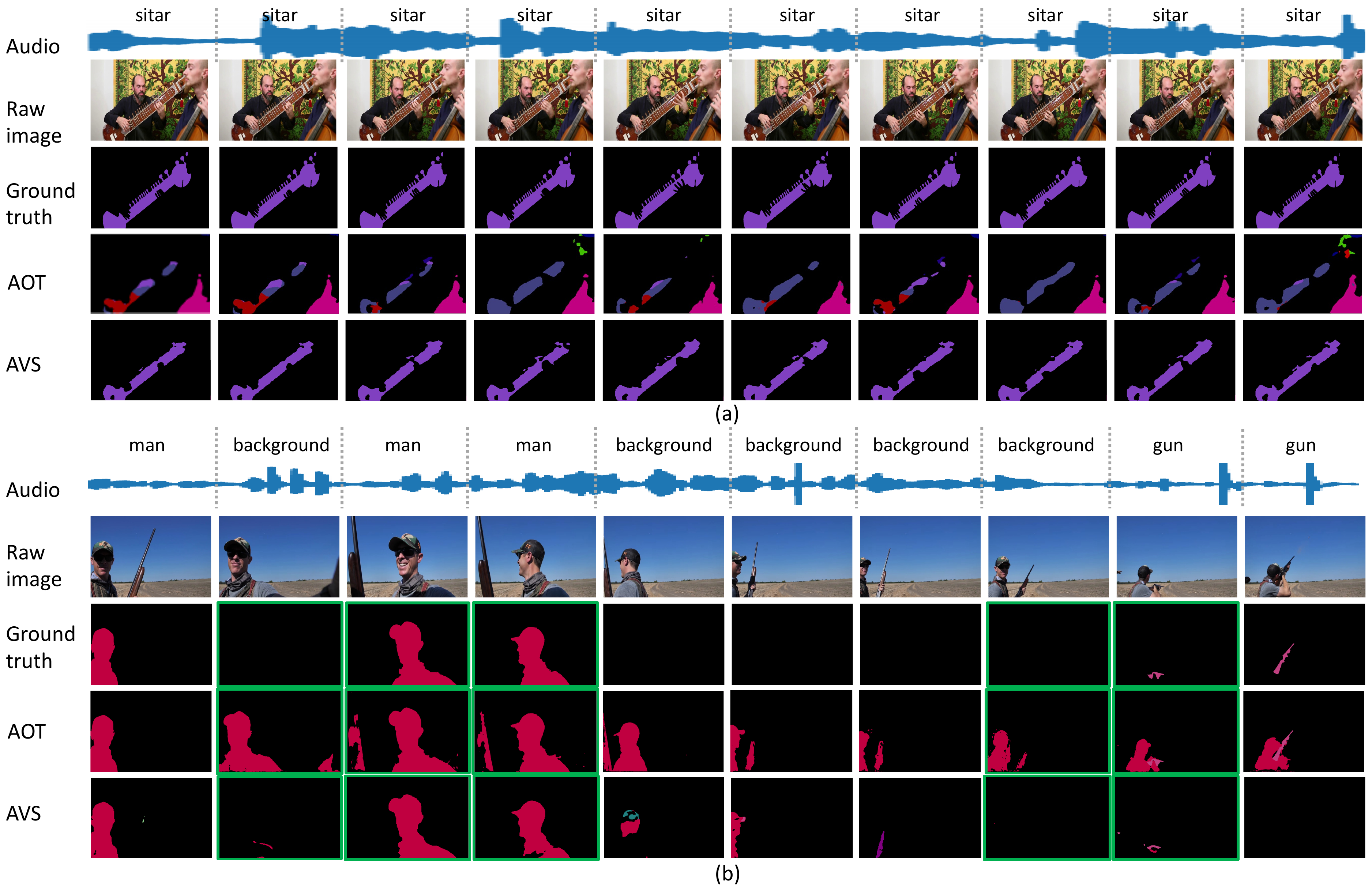} 
\vspace{-6mm}
\caption{\textbf{Qualitative examples of the VOS method AOT~\cite{yang2021aot} and our AVS method under the fully-supervised AVSS setting.} AVS model with audio guidance performs better to segment the correct audio-related objects and give more accurate semantic prediction.
}
\label{fig:example_avss_vos}
\end{figure*}

\noindent\textbf{Qualitative comparison between AVS and VOS.}
We also display some qualitative examples to compare the AOT method with our AVS model under the AVSS setting.
As shown in Fig.~\ref{fig:example_avss_vos}(a), the VOS method AOT segments the \emph{cello} at the lower right corner over the video frames which is actually not making sound and predict the \emph{guitar} with incorrect category. In contrast, our AVS model accurately segments the sounding \emph{guitar} with correct semantic.
In Fig.~\ref{fig:example_avss_vos}(b), when the sounding objects changes (green boxes), the AOT still segments both the \emph{man} and the \emph{gun} while our AVS model enables to merely segment the sounding one, \ie, the speaking \emph{man} in the third and fourth frames and the \emph{gun} in the last two frames.
These results again verify that audio information is helpful under the more challenging audio-visual semantic segmentation.

\begin{table}[t]
\caption{\textbf{Impact of audio signal and TPAVI.} Results (mIoU) of AVS model both with and without the TPAVI module.  The middle row indicates directly adding the audio and visual features, which already improves performance under the MS3 and the AVSS settings.  The TPAVI module further enhances the results over all settings and backbones. }
\small
\begin{center}
\begin{threeparttable}
\setlength{\tabcolsep}{3pt}
  \begin{tabular}{lp{0.6cm}<{\centering}p{1.0cm}<{\centering}p{0.6cm}<{\centering}p{1.0cm}<{\centering}p{0.6cm}<{\centering}p{1.0cm}<{\centering}}
   \toprule[0.8pt]
       \multirow{2}{*}{Method}  & \multicolumn{2}{c}{S4}  & \multicolumn{2}{c}{MS3} & \multicolumn{2}{c}{AVSS} \\ 
    \cmidrule(r){2-3}\cmidrule(r){4-5}\cmidrule(r){6-7} 
                  & Res50 & PVT-v2   & Res50 & PVT-v2  & Res50 & PVT-v2 \\ \midrule
    without TPAVI & 70.12    & 77.76  & 43.56    & 48.21 & 17.34 & 27.71 \\
    with A$\oplus$V &  70.54 & 77.65 & 45.69 & 51.55 & 19.85 & 28.94 \\
    with TPAVI & \textbf{72.79} & \textbf{78.74} &  \textbf{46.64} & \textbf{53.06} & \textbf{20.18}  & \textbf{29.77}\\
    \bottomrule[0.8pt]
   \end{tabular}
\end{threeparttable}
\end{center}\label{table:w_wo_tapvi}
\vspace{-4mm}
\end{table}

\begin{figure*}[t]
\centering
\includegraphics[width=\textwidth]{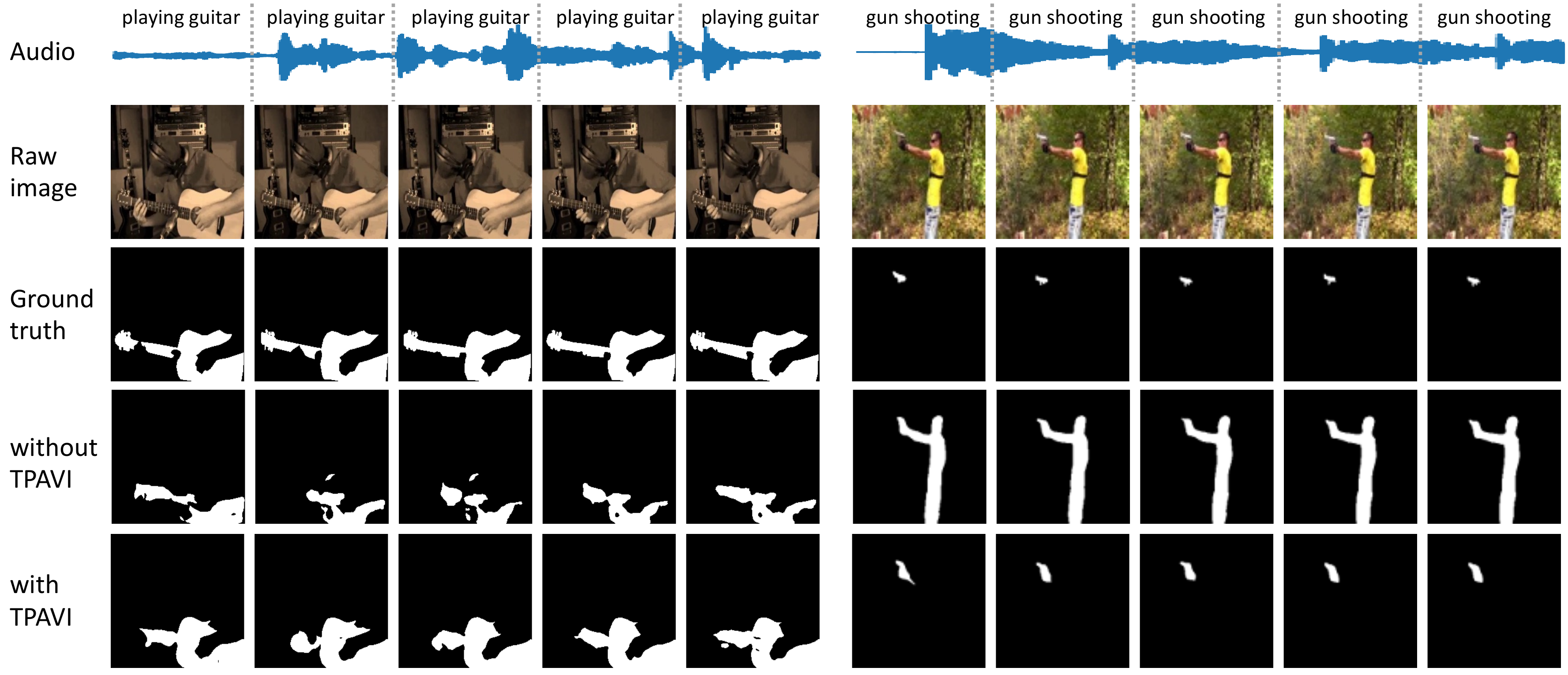}
\vspace{-6mm}
\caption{\textbf{Qualitative results under the semi-supervised S4 setting}. Predictions are generated by the ResNet50-based AVS model. Two benefits are noticed by introducing the audio signal (TPAVI): 1) learning the shape of the sounding object, \eg, \emph{guitar} in the video ({\sc Left}); 2) segmenting according to the correct sound source, \eg, the \emph{gun} rather than the \emph{man} ({\sc Right}).}
\label{fig:w_wo_tpavi_ssss}
\vspace{-4mm}
\end{figure*}

\begin{figure*}[t]
\centering
\includegraphics[width=\textwidth]{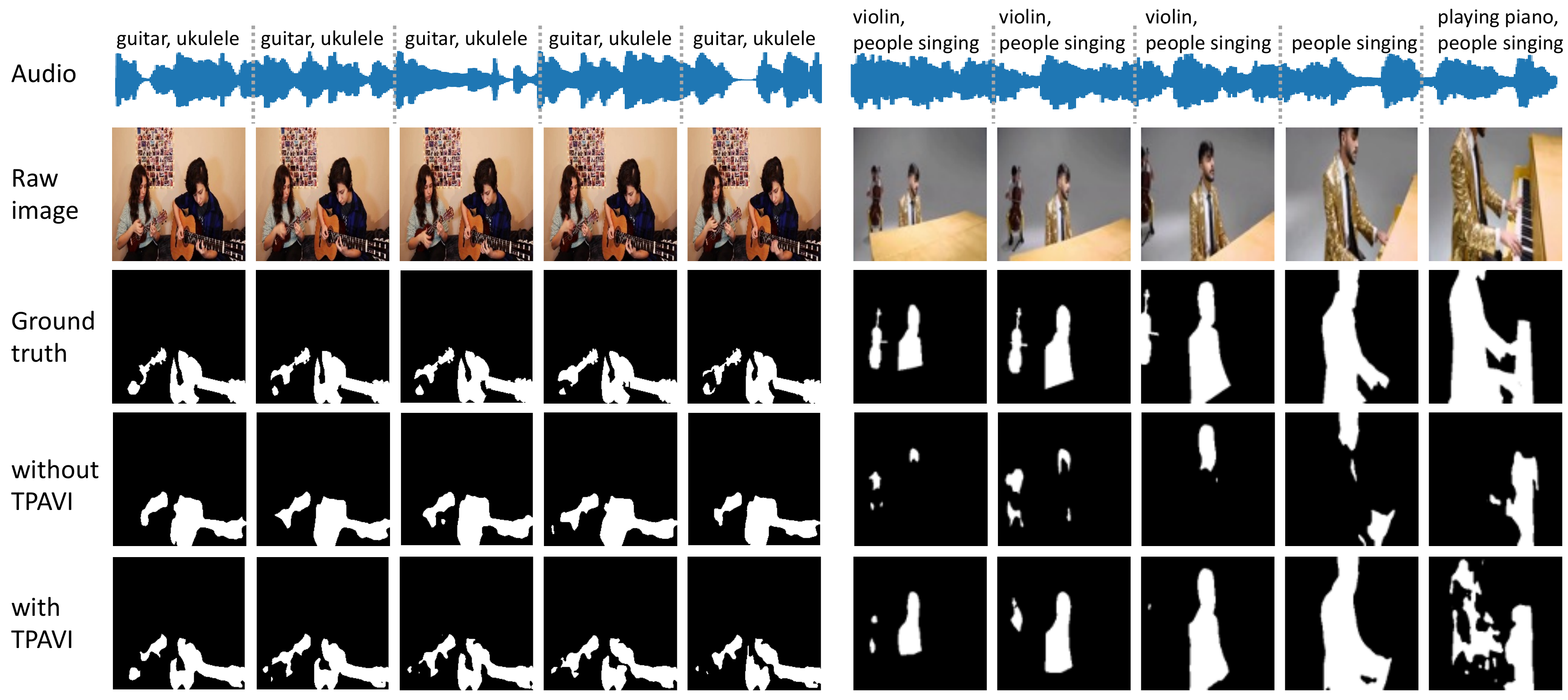}
\vspace{-6mm}
\caption{\textbf{Qualitative results under the fully-supervised MS3 setting}. The predictions are obtained by the PVT-v2 based AVS model. Note that AVS with TPAVI uses audio information to perform better in terms of 1) filtering out the distracting visual pixels that do not correspond to the audio, \ie, the \emph{human hands} ({\sc Left}); 2) segmenting the correct sound source in the visual frames that matches the audio more accurately, \ie, the \emph{singing person} ({\sc Right}).}
\label{fig:w_wo_tpavi_msss}
\vspace{-3mm}
\end{figure*}

\begin{figure*}[t]
\centering
\includegraphics[width=\textwidth]{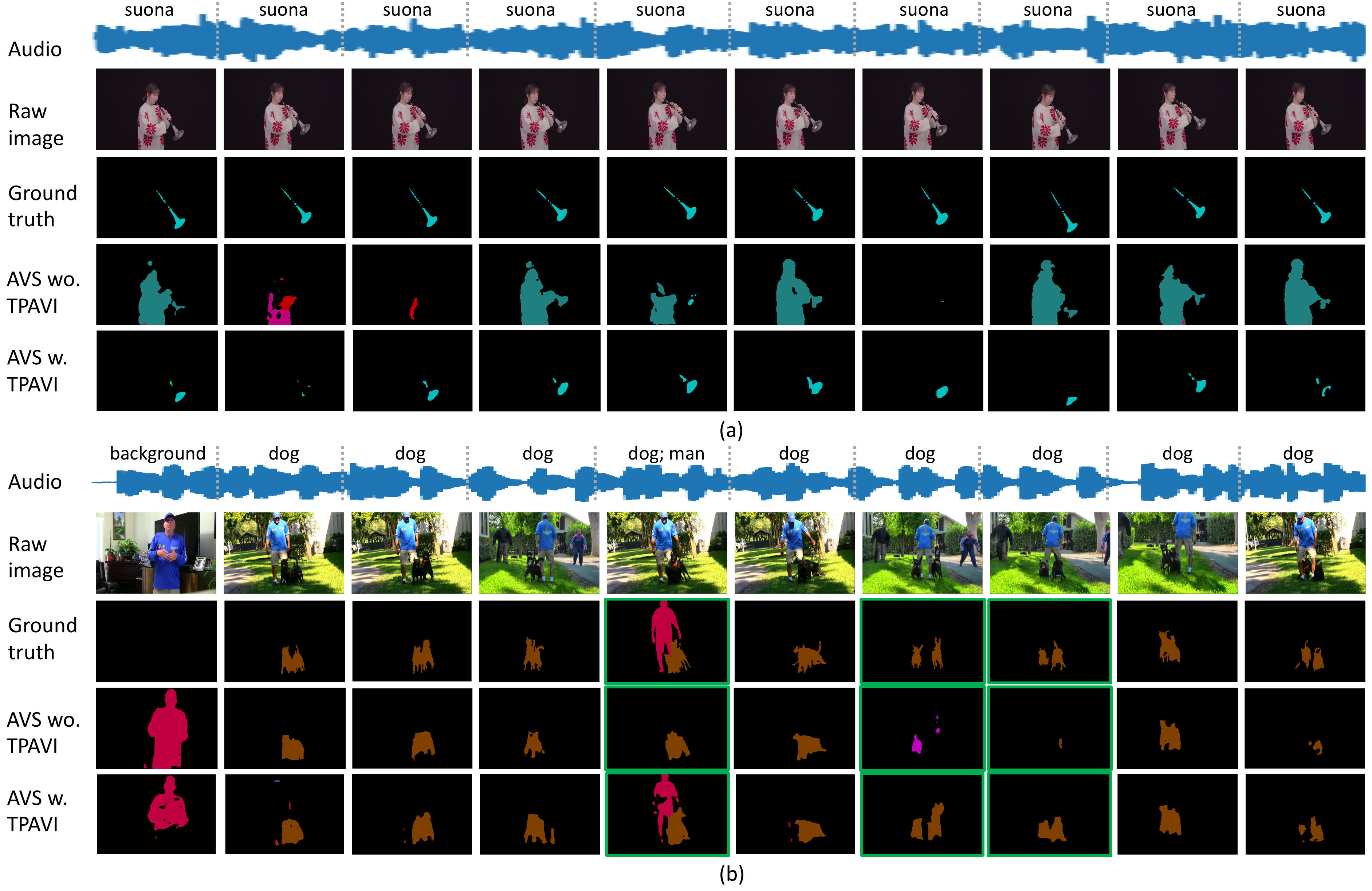}
\vspace{-6mm}
\caption{\textbf{Qualitative results under the fully-supervised AVSS setting.} The predictions are obtained by the PVT-v2 based AVS model. With the TPAVI module, the AVS model focuses on segmenting the objects which are making sounds, and with more complete shape and correct semantics.
}
\label{fig:w_wo_tpavi_avss}
\end{figure*}

\subsection{Model Analysis}\label{sec:exp_AVS_study}
\noindent\textbf{Impact of audio signal and TPAVI}.
As illustrated in Fig.~\ref{fig:TPAVI_module}, the TPAVI module is used to formulate the audio-visual interactions from a temporal and pixel-wise level, introducing the audio information to explore the visual segmentation.
%
We conduct an ablation study to explore its impact as shown in Table~\ref{table:w_wo_tapvi}.
Two rows show the proposed AVS method with or without the TPAVI module, while {``A$\oplus$V''} indicates directly adding the audio to visual features.
It will be noticed that adding the audio features to the visual ones does not result in a clear difference under the S4 setting, but lead to a distinct gain under the MS3 and AVSS settings.
This is consistent with our hypothesis that audio is especially beneficial to samples with multiple sound sources,
because the audio signals can guide which object(s) to segment.
Furthermore, with the power of our TPAVI module, we can achieve a temporal and pixel-wise mapping.
With TPAVI, each visual pixel hears the current sound and the sounds at other times, while simultaneously interacting with other pixels.
The physical interpretation is that the pixels with high similarity to the same sound are more likely to belong to one object.
TPAVI helps further enhance the performance over various settings and backbones, \eg, 72.79\% \textit{vs.} 70.54\% and 20.18\% \textit{vs.} 19.85\% when using ResNet50 as the backbone under the S4 and the AVSS settings, and 53.06\% \textit{vs.} 51.55\% if using PVT-v2 under the MS3 setting.

\begin{figure*}[t]
\centering
\includegraphics[width=\textwidth]{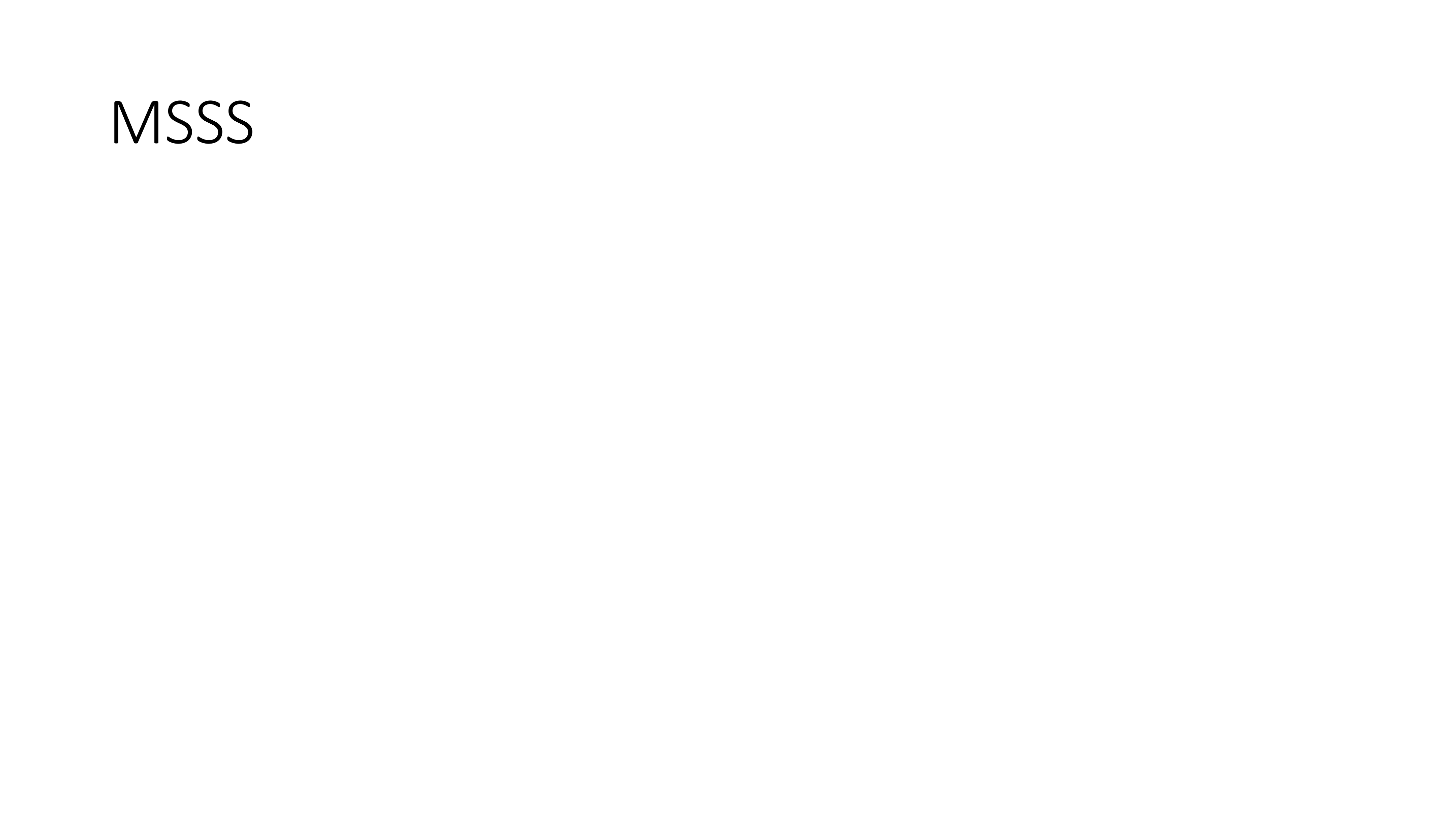}
\vspace{-6mm}
\caption{\textbf{Audio-visual attention maps that come from the fourth stage TPAVI.} 
Darker brown color indicates a higher response.
Such heatmaps are usually adopted as the final results for the SSL task, while they are just the intermediate output of the TPAVI module in our AVSS framework. These results reveal that the TPAVI helps the model focus more on the visual regions that are semantic-corresponding to the audio.}
\label{fig:audio_visual_attention}
\end{figure*}

We also visualize some qualitative examples to reflect the impact of TPAVI on AVS task under different settings. For the S4 setting, as shown in Fig.~\ref{fig:w_wo_tpavi_ssss}, the baseline method with TPAVI depicts the shape of sounding object better, \eg, the \emph{guitar} in the left video, while it can only segment several parts of the guitar without TPAVI. Such benefit can also be observed in the MS3 setting, as shown in Fig.~\ref{fig:w_wo_tpavi_msss}, the model enables to ignore those pixels of \emph{human hands} with TPAVI.
More importantly, with TPAVI, the model is able to segment the correct sounding object and ignore the potential sound sources which actually do not make sounds, \eg, the \emph{man} on the right of Fig.~\ref{fig:w_wo_tpavi_ssss}.
Also, the ``AVS w. TPAVI'' has a stronger ability to capture multiple sound sources. As shown on the right of Fig.~\ref{fig:w_wo_tpavi_msss}, the \emph{person} who is singing is mainly segmented with TPAVI but is almost lost without TPAVI. 
The impact of audio and TPAVI can also be verified under the AVSS setting. As shown in Fig.~\ref{fig:w_wo_tpavi_avss}a, ``AVS wo. TPAVI'' tends to segment the audio-unrelated part, \ie, the \emph{woman}. Besides, the sounding object \emph{suona} is not recognized in most of the video frames or recognized with incorrect semantics using ``AVS wo. TPAVI''. While the AVS model with TPAVI enables to focus on segmenting the truly sounding objects.
In Fig.~\ref{fig:w_wo_tpavi_avss}b, both ``AVS w. TPAVI'' and ``AVS wo. TPAVI'' incorrectly segments the silent \emph{man} at the initial frame.
The reason may be that the background noise misleads the model to give unnecessary predictions.
But ``AVS w. TPAVI'' successfully recognizes the speaking \emph{man} at the fifth frame and generates a more complete shape with more accurate semantics of the sounding \emph{dogs} in the subsequent frames (green boxes in the figure).
We argue that it is hard for a model without audio guidance to predict for AVS task because the model only learns to fit the provided ground truth and will not perceive the audio-visual correspondence.
These results show the advantages of utilizing the audio signals, which helps to segment more accurate audio-visual semantic-corresponding pixels.

Besides, we also visualize the audio-visual attention matrices to explore what happens in the cross-modal fusion process of TPAVI.
In detail, the attention matrix is obtained from $\alpha_i$ in Eq.~\eqref{eq:tpavi} of the fourth stage TPAVI.
We upsample it to have the same shape as the video frame. 
This is visually similar to the localization heatmap of these SSL methods, but only the intermediate result in our AVS method.
As shown in Fig.~\ref{fig:audio_visual_attention}, the high response area basically overlaps the region of sounding objects.
It suggests that TPAVI builds a mapping from the visual pixels to the audio signals, which is semantically consistent.

\begin{table}[t]
\caption{\textbf{Effectiveness of $\mathcal{L}_{\text{AVM}}$.} The two variants of $\mathcal{L}_{\text{AVM}}$ both bring a clear performance gain compared with only using a standard BCE loss.
}
\begin{center}
\begin{threeparttable}
\setlength{\tabcolsep}{6pt}
\begin{tabular}{lcccc}
   \toprule[0.8pt]
       \multirow{2}{*}{Objective function}  & \multicolumn{2}{c}{MS3 (mIoU)} & \multicolumn{2}{c}{AVSS (mIoU)}\\ 
    \cmidrule(r){2-3}\cmidrule(r){4-5}
                  & ResNet50 & PVT-v2  & ResNet50 & PVT-v2 \\ \midrule
    $\mathcal{L}_{\text{BCE}}$  & 46.64    & 53.06 & 18.88 & 29.17 \\
    $\mathcal{L}_{\text{BCE}}$ + $\mathcal{L}_{\text{AVM-VV}}$ & 46.71 & 53.77 & 19.65 & 29.62 \\
    $\mathcal{L}_{\text{BCE}}$ + $\mathcal{L}_{\text{AVM-AV}}$ & \textbf{47.88} & \textbf{54.00} & \textbf{20.18} & \textbf{29.77} \\
    \bottomrule[0.8pt]
   \end{tabular}
\end{threeparttable}
\end{center}\label{table:loss_avm_study}
\end{table}

\noindent\textbf{Effectiveness of $\mathcal{L}_\text{AVM}$}. We expect that constructing the mapping between audio and visual features will enhance the network's ability to identify the correct objects. 
Therefore, we propose a $\mathcal{L}_\text{AVM}$ loss to introduce a soft constraint for training.
We only apply $\mathcal{L}_\text{AVM}$ in the fully-supervised MS3 setting and AVSS setting because the change of sounding objects only happens there.

\begin{figure*}[t]
\centering
\includegraphics[width=\textwidth]{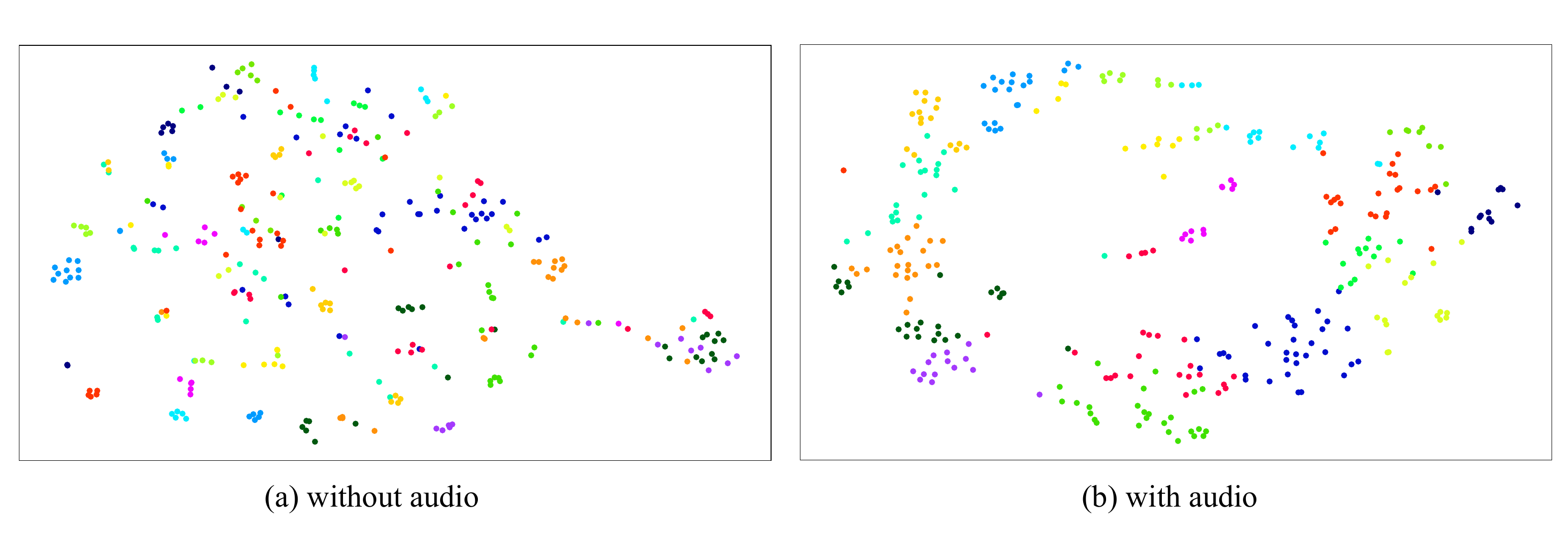}
\vspace{-9mm}
\caption{\textbf{T-SNE~\cite{tsne} visualization of the visual features, trained with or without audio.} These results are from the test split of the Multi-sources subset. We first use principal component analysis (PCA) to divide the audio features into $K=20$ clusters. Then we assign the audio cluster labels to the corresponding visual features and conduct t-SNE visualization. The points with the same color share the same audio cluster labels. It can be seen that when training is accompanied by audio signals (right), the visual features illustrate a closer trend with the audio feature distribution, \ie, points with the same colors gather together, which indicates an audio-visual correlation has been learned.  (Best viewed in color.) }
\label{fig:tsne}
\end{figure*}

As shown in Table~\ref{table:loss_avm_study}, 
we explore two variants of the $\mathcal{L}_\text{AVM}$ loss.
$\mathcal{L}_{\text{AVM-AV}}$ is the one introduced in Eq.~\eqref{eq:avmloss}.
It encourages the visual features masked by the segmentation result to be consistent with the corresponding audio features in a statistical way, \ie, both depicting the sounding objects. 
%
Alternatively, $\mathcal{L}_{\text{AVM-VV}}$ first finds the closest audio partner for each candidate audio, and then computes the KL distance of the corresponding visual features (also masked by the segmentation results).
This is based on the idea that if two clips share similar audio signals, the visual features of their sounding objects should also be similar.
As shown in Table~\ref{table:loss_avm_study}, both variants achieve a clear performance gain.
%
For example, $\mathcal{L}_{\text{AVM-AV}}$ improves the mIoU by around 1\% under the MS3 and AVSS settings.
%
This demonstrates the benefits of introducing such an audio-visual constraint.
We use $\mathcal{L}_{\text{AVM-AV}}$,
since  $\mathcal{L}_{\text{AVM-VV}}$ inconveniently requires a ranking operation.

\begin{table}[t]
\caption{\textbf{Cross-modal fusion at various stages, measured by mIoU (\%).} In all the settings, the model achieves the best performance when the TPAVI module is used in all four stages.}
\begin{center}
\begin{threeparttable}
    \begin{tabular}{llp{0.4cm}<{\centering}p{0.4cm}<{\centering}p{0.4cm}<{\centering}p{0.5cm}<{\centering}|p{0.4cm}<{\centering}p{0.4cm}<{\centering}p{0.5cm}<{\centering}}
  \toprule[0.8pt]
      \multirow{2}{*}{Setting}  & \multirow{2}{*}{Backbone}  & \multicolumn{7}{c}{$i$-th stage of Encoder, $i\in\{1,2,3,4\}$} \\ 
    \cmidrule(r){3-9} 
      & & 1 & 2 & 3 & 4 & 3,4 & 2,3,4 & 1,2,3,4 \\ \midrule
    \multirow{2}{*}{S4} & ResNet50 & 68.55 & 69.56 & \textbf{71.30} & 69.99 & 71.29 & 71.98 & \textbf{72.79} \\ 
    & PVT-v2 & 78.30 & \textbf{78.58} & 78.02 & 77.70 & 78.19 & 78.47 & \textbf{78.74} \\ \midrule
    \multirow{2}{*}{MS3} & ResNet50 & 41.62 & 42.37 & \textbf{43.02} & 42.29 & 44.84 & 45.98 & \textbf{47.88} \\
    & PVT-v2 & 46.16 & 48.79 & 47.35 & \textbf{49.01} & 49.79 & 50.53 & \textbf{54.00} \\ \midrule
    \multirow{2}{*}{AVSS} & ResNet50 & \textbf{19.29} & 18.39 & 18.89 & 17.96 & 18.16 & 18.44 & \textbf{20.18} \\
    & PVT-v2 & 28.62 & \textbf{29.19} & 29.07 & 28.59 & 28.78 & 28.73 & \textbf{29.77} \\
    \bottomrule[0.8pt]
  \end{tabular}
\end{threeparttable}
\end{center}\label{table:tpavi_at_different_stages}
\vspace{-2mm}
\end{table}

\noindent\textbf{Cross-modal fusion at various stages}.
The TPAVI module is a plug-in architecture that can be applied in any stage for cross-modal fusion.
As shown in Table~\ref{table:tpavi_at_different_stages}, when the TPAVI module is used in different single stages, the segmentation performance fluctuates.
{{For the variant based on the ResNet50 backbone, the model achieves the best performance when employing the TPAVI module at the third stage under both S4 and MS3 settings and at the first stage under AVSS setting. 
As for the PVT-v2 based model, it is better to use the TPAVI module at the second stage in the S4 and AVSS settings and at the fourth stage under MS3 setting.
%
The AVSS setting needs to further predict the semantic label for each pixel and thus may benefit more from the early stage having a large receptive field.}}
Since our decoder architecture adopts a skip-connection, it would be beneficial to apply the TPAVI modules in multiple stages, as verified in the right part of Table~\ref{table:tpavi_at_different_stages}.
For example, under the MS3 setting, applying TPAVI at all four stages would increase the metric mIoU from  $49.01\%$ to $54.00\%$, with a gain of $4.99\%$.
%
It indicates the model has the ability to fuse and balance the features from multiple stages.

\begin{table}[t]
\caption{\textbf{Performance with different initialization strategies under the MS3 setting.}
Compared to training from scratch under the MS3 setting, we observe a significant performance improvement if pre-training the model on the Single-source subset. Note the proposed $\mathcal{L}_{\text{AVM}}$ loss is used in all the experiments of the Table. The metric is mIoU.}
\begin{center}
\begin{threeparttable}
\begin{tabular}{lcccc}
  \toprule[0.8pt]
      \multirow{2}{*}{Method}  & \multicolumn{2}{c}{ From scratch}  & \multicolumn{2}{c}{ Pretrained on Single-source} \\ 
    \cmidrule(r){2-3}\cmidrule(r){4-5} 
                  & ResNet50 & PVT-v2   & ResNet50 & PVT-v2 \\ \midrule
    wo. TPAVI & 43.56    & 48.21 & \textbf{45.50}    & \textbf{50.59} \\
    w. TPAVI  & 47.88    & 54.00 & \textbf{54.33}    & \textbf{57.34} \\
    \bottomrule[0.8pt]
  \end{tabular}
\end{threeparttable}
\end{center}\label{table:load_ssss_for_msss}
\end{table}

\begin{figure*}[h]
\centering
\includegraphics[width=\textwidth]{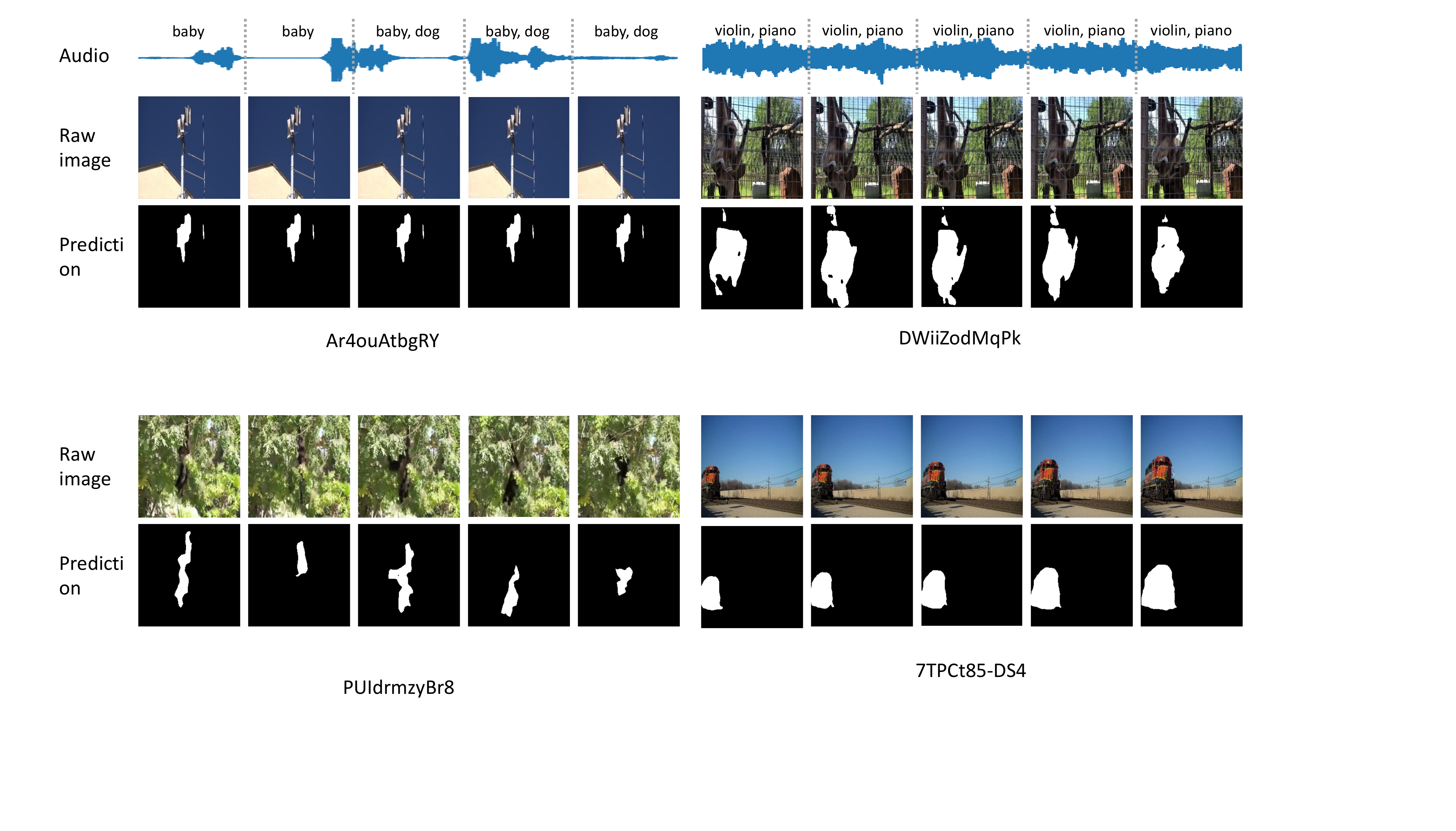}
\vspace{-6mm}
\caption{\textbf{Qualitative examples of applying the pretrained AVS model {under the MS3 setting} to unseen videos.} The caption in each sub-figure indicates the sounding object(s) accordingly. There are almost no videos having the same category as these sounding objects during AVS model training. The pretrained AVS model gains the ability to segment the correct sounding object(s) in both single and multi sources.}
\label{fig:unseen_objects_inference}
\end{figure*}

\noindent\textbf{Pre-training on the Single-source subset.}
As introduced in Sec.~\ref{sec:dataset} of the paper, the videos in the Multi-sources subset share similar categories to those in the Single-source subset.
A natural idea is whether we can pre-train the model on the Single-source subset to help deal with the MS3 problem.
As shown in Table~\ref{table:load_ssss_for_msss}, we test two initialization strategies, \ie, from scratch or pretrained on the Single-source subset.
It is verified that the pre-training strategy is beneficial in all the settings, whether we use the audio information (``w. TPAVI'') or not (``wo. TPAVI'').
Taking the PVT-v2 based AVS model for example, the mIoU is improved from $48.21\%$ to $50.59\%$ (by $2.38\%$) and from $54.00\%$ to $57.34\%$ (by $3.34\%$), respectively without or with TPAVI.
%
%
The phenomenon is more obvious if using ResNet50 as the backbone and adopting the TPAVI module, where the mIoU increases from $47.88\%$ to $54.33\%$ (by $6.45\%$).
With pre-training on the Single-source subset, the model can learn prior knowledge about the audio-visual correspondence, \ie, the matching relationship between the visual objects and sounds.
This kind of knowledge is naturally beneficial.
%

\noindent\textbf{T-SNE visualization analysis.}
We also visualize the visual features with or without TPAVI module to analyze whether the network has built a connection between the audio and the visual features. 
Specifically, on the test split of the Multi-sources set, we use the PVT-v2 based AVS model to obtain the visual features.
Since the Multi-source set do not have category labels (its videos may contain several categories), 
we use the principal component analysis (PCA) to divide the audio features into $K=20$ clusters. Then we assign the audio cluster labels to the corresponding visual features. In this case, if the audio and the visual features are correlated, the visual features should be clustered as well. We use the t-SNE visualization to verify this assumption.  
As shown in Fig.~\ref{fig:tsne}a, without audio signals, the learned visual features distribute chaotically;
whereas in Fig.~\ref{fig:tsne}b, the visual features sharing the same audio labels tend to gather together.
This indicates that the distribution of the visual features and  audio features are highly correlated.

\noindent\textbf{Segmenting unseen objects.} We restrict the study under the MS3 setting as it does not need the model to predict the actual category labels for unseen objects but still requires the model to predict the sounding objects.
We display some qualitative visualizations on real-world videos whereas the category of sounding objects are barely not appeared in the training set of AVS model.
As shown in Fig.~\ref{fig:unseen_objects_inference}, the pretrained AVS model has a certain ability to segment the correct sounding objects in the case of a single sound source (a), multiple visible objects (b, c), and multiple sound sources (d).
We speculate that the pretrained AVS model learned some prior knowledge about audio-visual correspondence from the AVSBench dataset that helps it generalize to even unseen videos and give possibly accurate pixel-level segmentation.

\section{Conclusion}
We explore the task of audio-visual segmentation (AVS), which aims to generate pixel-level segmentation masks for sounding objects in audible videos.
To facilitate research on AVS, we build and enrich the audio-visual segmentation benchmark (AVSBench) that contains the single-source, multi-sources and semantic-labels subsets.
Accordingly, three task settings are explored: the semi-supervised single-source AVS (S4), fully-supervised multi-source AVS (MS3) and the fully-supervised audio-visual semantic segmentation (AVSS).
We presented a new pixel-level method to serve as a strong baseline and work for those three settings, which includes a TPAVI module to encode the pixel-wise audio-visual interactions within temporal video sequences and a regularization loss that is designed to help the model learn audio-visual correlations. We compared our method with several existing state-of-the-art methods from related tasks on AVSBench, and further demonstrated that our method can build a connection between the sound and the appearance of an object. For future work, we will create a large-scale synthetic dataset for model pre-training.

\ifCLASSOPTIONcaptionsoff
  \newpage
\fi

\bibliographystyle{IEEEtran}
\bibliography{IEEEabrv,sample-base}

\end{document}